\documentclass[runningheads]{llncs}

\usepackage[mobile]{eccv}

\usepackage{eccvabbrv}

\usepackage{graphicx}
\usepackage{booktabs}
\usepackage{makecell}

\usepackage[accsupp]{axessibility}  

\usepackage{hyperref}

\usepackage{orcidlink}

\renewcommand{\subsubsection}[1]{\smallskip\noindent\textbf{#1}}

\usepackage[table]{xcolor}
\usepackage{multicol}
\usepackage{multirow}
\usepackage{flushend}
\usepackage{todonotes}
\usepackage{array}
\usepackage{tabularray}
\usepackage{footnote}

\definecolor{lightgray}{rgb}{0.85,0.85,0.85}

\usepackage{booktabs,arydshln}

\makeatletter
\def\adl@drawiv#1#2#3{%
        \hskip.5\tabcolsep
        \xleaders#3{#2.5\@tempdimb #1{1}#2.5\@tempdimb}%
                #2\z@ plus1fil minus1fil\relax
        \hskip.5\tabcolsep}
\newcommand{\cdashlinelr}[1]{%
  \noalign{\vskip\aboverulesep
           \global\let\@dashdrawstore\adl@draw
           \global\let\adl@draw\adl@drawiv}
  \cdashline{#1}
  \noalign{\global\let\adl@draw\@dashdrawstore
           \vskip\belowrulesep}}
\makeatother

\begin{document}

\title{FreeFlow: A Bias-free Hierarchical \\Transformer for Optical Flow Estimation} 

\titlerunning{FreeFlow: A Bias-free Hierarchical Transformer for Optical Flow Estimation}

\author{Vladislav Bargatin\inst{1,2\footnotemark[1]}\orcidlink{0009-0007-2621-1577} \and
Alexander Yakovenko\inst{1,2,3\footnotemark[1]\footnotemark[4]}\orcidlink{0000-0003-3105-512X} \and \\
Khaled Abud\inst{1,2,3}\orcidlink{0009-0009-7131-5839} \and
Dmitriy Vatolin\inst{3}\orcidlink{0000-0002-8893-9340}
}

\authorrunning{V.~Bargatin et al.}

\institute{AI Center, Lomonosov MSU, Moscow, Russia \and
Lomonosov Moscow State University, Moscow, Russia \and
MSU Institute for Artificial Intelligence, Moscow, Russia\\
\email{\{vladislav.bargatin, alexander.yakovenko\}@graphics.cs.msu.ru}, \email{\{khaled.abud, dmitriy\}@graphics.cs.msu.ru}\\
\url{https://github.com/msu-video-group/freeflow}}

\maketitle

\renewcommand{\thefootnote}{\fnsymbol{footnote}}
\footnotetext[1]{Equal contribution}
\footnotetext[4]{Corresponding author}

\begin{abstract}
  Optical flow methods typically rely on task-specific inductive biases, such as correlation volumes, feature warping, and iterative refinement, among others, to reach high accuracy. While effective, such biases constrain the model to predefined heuristics, which can limit its expressivity and lead to more complex pipelines and additional computational cost. We present FreeFlow, a hierarchical transformer built without any flow-specific components, using instead a single feed-forward encoder--decoder. FreeFlow combines three attention variants: window attention for local processing, shifted-window attention for cross-window information exchange, and a global attention operating at a reduced resolution. The resulting architecture scales naturally with model capacity, enabling a consistent accuracy gain from small to large variants. Despite the absence of standard inductive biases, FreeFlow achieves state-of-the-art results on major benchmarks, including Sintel (0.68/1.48 EPE on Clean/Final), KITTI-2015 (3.23 Fl-all), and Spring (3.192 1px), while remaining memory efficient at 1080p inference.
\end{abstract}
\section{Introduction}
\label{sec:intro}

Optical flow estimation (the dense per-pixel motion between frames) is a fundamental task in low-level vision, with applications ranging from video understanding\cite{piergiovanni2019representation, sun2018optical, zhao2020improved} and tracking to video restoration and synthesis\cite{huang2022real, liu2020video, xu2019quadratic, chan2021basicvsr}.

Early optical flow methods posed estimation as variational optimization\cite{lucas1981iterative, horn1981determining}, later accumulating stronger regularization \cite{sun2010secrets}, coarse-to-fine schemes, and hand-crafted descriptors\cite{weinzaepfel2013deepflow} at the cost of growing complexity.

Deep learning initially simplified optical flow, as FlowNet\cite{dosovitskiy2015flownet} showed that a feed-forward network can regress flow directly, but later work reintroduced classical biases in highly effective yet hard-wired architectures. PWC-Net\cite{sun2018pwc} combined pyramids, warping, and local correlation volumes, while RAFT\cite{teed2020raft} popularized iterative refinement over an all-pairs correlation volume and convex upsampling; many follow-ups then build upon these foundations with additional modules for occlusions\cite{jiang2021learning_GMA}, temporal cues\cite{shi2023videoflow, dong2024memflow}, and training/inference refinements\cite{wang2025sea}. Which improve accuracy, but lead to complex pipelines that are harder to modify, scale, and repurpose beyond optical flow.

In parallel, the broader computer vision literature has moved in the opposite direction: vision transformers\cite{dosovitskiy2020image} increasingly replace bespoke pipelines in detection\cite{carion2020end}, segmentation\cite{kirillov2023segment, kerssies2025your}, depth prediction\cite{yang2024depth, yang2024depth2}, and 3D tasks\cite{wang2024dust3r, wang2025vggt, jin2025lvsm}, driven by the observation that generic, data-driven feed-forward models can learn the required structure from scale and supervision. This trend motivates revisiting optical flow through the same lens: \textit{can we omit flow-specific components and still obtain a state-of-the-art approach?}

Several recent approaches move toward more generic architectures, but still retain flow-specific structure or incur practical limitations. DDVM\cite{saxena2023surprising} casts optical flow prediction in a diffusion framework, but the prediction is still produced through iterative process. CroCo\cite{weinzaepfel2023croco} is close to a pure transformer, yet it operates at a relatively small fixed resolution and typically requires tiling for higher-resolution inputs, while still relying on a large convolutional decoder. Most recently, WAFT\cite{wang2025waft} and GeoViT\cite{wu2025study} explicitly aim for generality, but still rely on iterations and require warping (of features and input frames, respectively). Additionally, these approaches are trained at relatively small, sub-megapixel resolutions, which might become a limiting factor for accuracy at high-resolution inference~\cite{Bargatin_2025_ICCV}.

\begin{figure}[t]
    \begin{subfigure}[t]{0.49\linewidth}
        \centering
        \includegraphics[width=\linewidth]{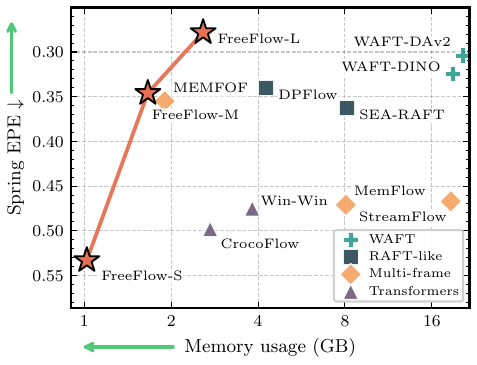}
        \caption{Spring EPE ($\downarrow$) versus 1080p inference memory (GB), with FreeFlow variants (S/M/L) illustrating model scaling.}    
        \label{fig:intro:scatter}
    \end{subfigure}
    \hfill
    \begin{subfigure}[t]{0.49\linewidth}
        \centering
        \includegraphics[width=\linewidth]{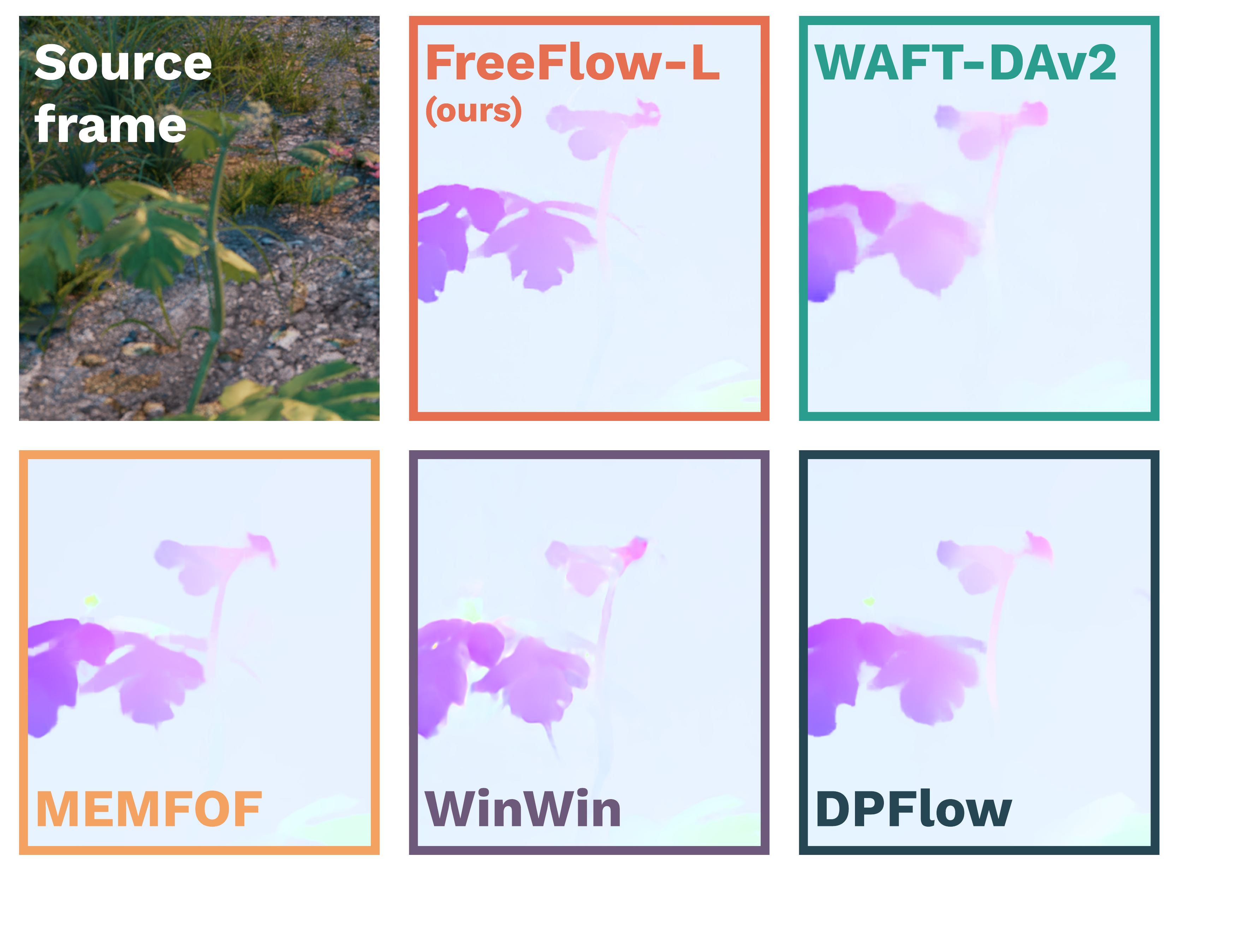}
        \caption{Qualitative comparison of predicted flow fields on a crop of a visually challenging scene with heavy blur (best viewed zoomed in).}
    \end{subfigure}
    \caption{\textbf{FreeFlow overview.} Our method achieves state-of-the-art accuracy on Spring with low 1080p inference memory and the sharpest motion borders among strong baselines, as evidenced by (a) the accuracy--memory scatter plot and (b) the qualitative crop gallery.}
    \label{fig:intro}
\end{figure}

We therefore propose FreeFlow, a bias-free hierarchical transformer for optical flow estimation. FreeFlow is designed for high-resolution processing, which recent work has shown to be beneficial for optical flow\cite{Bargatin_2025_ICCV} and depth estimation\cite{Bochkovskii2024:arxiv}. At high resolution, accurate flow requires both strong local reasoning (to preserve fine structures and motion boundaries) and effective long-range information flow (to resolve large displacements). FreeFlow addresses this with a hierarchical attention design that combines tiled processing with repeated local--global feature interaction: local attention focuses on within-region detail, cross-region exchange propagates information between neighboring tiles, and global mixing enables long-range correspondence. FreeFlow sets a new state-of-the-art on Sintel\cite{butler2012naturalistic} (EPE clean: 0.68; EPE final: 1.48), KITTI-2015\cite{Menze2015CVPR} (Fl-all: 3.23), and Spring\cite{mehl2023spring} (1px: 3.192).

Our key contributions are:
\begin{itemize}
	\item \textbf{Bias-free optical flow transformer.} We introduce FreeFlow, a hierarchical transformer for optical flow built \emph{without common flow-specific inductive biases and modules} (e.g., explicit cost/correlation volumes, warping-based update pipelines, and specialized upsampling), using a single end-to-end trainable architecture.
	\item \textbf{Hierarchical local–global feature interaction for high-resolution processing.} We propose a tiled transformer design with repeated local and long-range feature interaction, enabling accurate flow at high resolutions by combining local detail modeling with global information flow.
	\item \textbf{State-of-the-art performance across benchmarks.} FreeFlow achieves state-of-the-art results on Sintel, KITTI-2015, and Spring, all while being memory-efficient and scalable to smaller parameter counts.
\end{itemize}
\section{Related Works}

\subsection{Inductive Biases of Optical Flow}
Early optical flow methods formulated estimation as variational optimization under photometric constancy and smoothness regularization\cite{lucas1981iterative,horn1981determining,sun2010secrets,weinzaepfel2013deepflow}. Learning-based approaches later treated optical flow as supervised dense prediction\cite{dosovitskiy2015flownet}, and subsequent advances largely introduced explicit architectural priors tailored to correspondence estimation. Representative flow-specific inductive biases include multi-scale pyramids\cite{sun2018pwc, morimitsu2025dpflow}, warping-based alignment\cite{sun2018pwc, wang2025waft, wu2025study}, explicit correlation volumes for large-displacement matching\cite{teed2020raft, Bargatin_2025_ICCV}, iterative refinement, and convex upsampling\cite{teed2020raft, wang2025sea, Bargatin_2025_ICCV}. While these choices are effective, they have contributed to increasingly complex pipelines; recent work has started to remove individual priors\cite{kiefhaber2025removing} and move toward more generic formulations\cite{saxena2023surprising}, but existing methods typically retain some of these biases rather than eliminating them entirely.

\begin{table}[t]
    \caption{\textbf{Architectural Inductive Biases in Optical Flow Methods.} We summarize common design choices used to inject task-specific structure compared to our approach, which has no flow-specific inductive biases and uses a simple decoder head. 
    "DPT head" refers to the decoder head from Dense Prediction Transformer (DPT)~\cite{ranftl2021vision}.
    }
    \label{tab:ind-bias}
    \centering
    \newcommand{\yay}{\cellcolor{red!25}$\checkmark$\xspace}
\newcommand{\nay}{\cellcolor{green!25}$\times$\xspace}
\newcommand{\nayw}{\cellcolor{white!25}$\times$\xspace}

\newcommand{\rotangle}{0}

\resizebox{1.0\linewidth}{!}{
\begin{tabular}{l c c c c c c c c}
\toprule
\multirow{2}{*}{Method} &
\multirow{2}{*}{\rotatebox{\rotangle}{\shortstack[c]{Correlation \\ Volume}}} &
\multirow{2}{*}{\rotatebox{\rotangle}{\shortstack[c]{Feature \\ Warping}}} &
\multirow{2}{*}{\rotatebox{\rotangle}{\shortstack[c]{Pyramid \\ Refinement}}} &
\multirow{2}{*}{\rotatebox{\rotangle}{\shortstack[c]{Iterative \\ Refinement}}} &
\multirow{2}{*}{\rotatebox{\rotangle}{\shortstack[c]{Convex \\ Upsampling}}} &
\multirow{2}{*}{\rotatebox{\rotangle}{\shortstack[c]{Inference time \\ Tiling}}} &
\multirow{2}{*}{\rotatebox{\rotangle}{\shortstack[c]{Flow \\ Head}}} \\
\\
\midrule
PWC-Net~\cite{sun2018pwc} & \yay & \yay & \yay & \nay & \nay & \nay & \nay \\
RAFT~\cite{teed2020raft} & \yay & \nay & \nay & \yay & \yay & \nay & \nay \\
FlowFormer~\cite{huang2022flowformer} & \yay & \nay & \nay & \yay & \yay & \yay & \nay \\
UniMatch~\cite{xu2023unifying}& \nay & \yay & \yay & \yay & \yay & \nay & \nay \\
TransFlow~\cite{lu2023transflow} & \yay & \nay & \nay & \yay & \yay & \nay & \nay \\
DPFlow~\cite{morimitsu2025dpflow} & \yay & \nay & \yay & \yay & \yay & \nay & \nay \\
MEMFOF~\cite{Bargatin_2025_ICCV} & \yay & \nay & \nay & \yay & \yay & \nay & \nay \\
WAFT~\cite{wang2025waft} & \nay & \yay & \nay & \yay & \yay & \nay & \nay \\
Geo-VIT~\cite{wu2025study} & \nay & \yay & \nay & \yay & \yay & \yay & \nay \\
CroCo-Flow~\cite{weinzaepfel2023croco} & \nay & \nay & \nay & \nay & \nay & \yay & \cellcolor{yellow!25}DPT head\\
Win-Win~\cite{leroy2023win} & \nay & \nay & \nay & \nay & \nay & \nay & \cellcolor{yellow!25}DPT head\\
FreeFlow (ours) & \nay & \nay & \nay & \nay & \nay & \nay & \cellcolor{green!25}Simple Conv.\\
\bottomrule
\end{tabular}
}
\end{table}

\subsection{Vision Transformers in Dense Prediction}
Vision transformers\cite{dosovitskiy2020image} (ViTs) are now widely used for dense prediction and geometry-oriented tasks, enabled by large-scale data and a general architecture that can learn the dependencies from supervision rather than relying on hand-crafted priors. This has led to strong results across depth\cite{ranftl2021vision, yang2024depth, yang2024depth2, Bochkovskii2024:arxiv}, segmentation\cite{kirillov2023segment, kerssies2025your}, and 3D settings\cite{wang2024dust3r, wang2025vggt, jin2025lvsm}, where feed-forward transformer backbones provide a common foundation for dense outputs and geometric reasoning. Importantly, many of these systems remain architecturally simple. A common pattern is a ViT backbone combined with a convolutional prediction head or decoder for producing dense maps\cite{ranftl2021vision, ravisam}. Other approaches further reduce decoder structure and rely on minimal output token processing, suggesting that heavy convolutional decoders are not necessary to obtain competitive dense predictions \cite{kerssies2025your, jin2025lvsm}. However, scaling such models to high resolution is challenging; common workarounds such as downsampling or tiling can harm accuracy and limit long-range interaction. Swin\cite{liu2021swin, liang2021swinir} addresses the issue by using local-window self-attention and shifted windows to pass information across window boundaries, while Hiera\cite{ryali2023hiera} provides a simple hierarchical multi-scale ViT backbone for large images. Alternatively, DepthPro does high-resolution processing via a multi-scale pyramid design with late feature fusion\cite{Bochkovskii2024:arxiv}.

\subsection{Transformers in Optical Flow Estimation}
Recent transformer-based optical flow methods differ mainly in which flow-specific inductive biases they retain. Some methods keep explicit matching as a central operation via correlation/cost-based similarity: UniMatch and TransFlow follow this direction \cite{xu2023unifying, lu2023transflow}, while FlowFormer explicitly constructs and processes a 4D cost volume with a transformer-style decoder \cite{huang2022flowformer}.

While others move closer to generic transformer formulations, they unfortunately still leave some biases in place. CroCo-Flow was among the first transformer approaches with competitive accuracy, leveraging binocular pretraining for dense matching, but it relies on slow dense tiling at high resolution \cite{weinzaepfel2023croco}. Win-Win builds on CroCo to enable FullHD training and inference without tiling, but does not improve over CroCo-Flow in accuracy \cite{leroy2023win}. WAFT and GeoViT remove cost volumes but retain iterative warping-based updates (warping features and input images, respectively) \cite{wang2025waft, wu2025study}. Overall, transformers have often been incorporated by incrementally replacing parts of established optical flow pipelines, which can improve performance yet further diversify and complicate the set of design choices. We summarize these choices in \cref{tab:ind-bias} using common inductive biases and compare to our bias-free approach.

\section{Method}
\begin{figure*}[t]
\centering
\includegraphics[width=\textwidth]{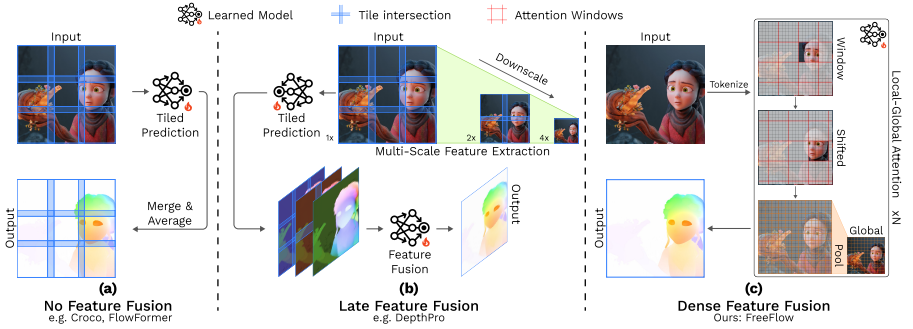}
\caption{\textbf{Comparison of high-resolution prediction techniques.} (a) \emph{No Feature Fusion:} per-tile prediction without inter-tile feature exchange (limited global coherence; long-range motions spanning multiple tiles not captured). (b) \emph{Late Feature Fusion:} multi-scale features are fused only in the decoder (global context arrives late and may not propagate to full-resolution features; with small overlap and no information flow across tile borders, seams may remain visible; see supplementary). (c) \emph{Dense Feature Fusion (ours):} repeated cross-tile/global feature exchange across the network (information flow is not restricted to the decoder).}
\label{fig:tiling-comp}
\end{figure*}

\begin{figure*}[t]
\centering
\includegraphics[width=.99\textwidth]{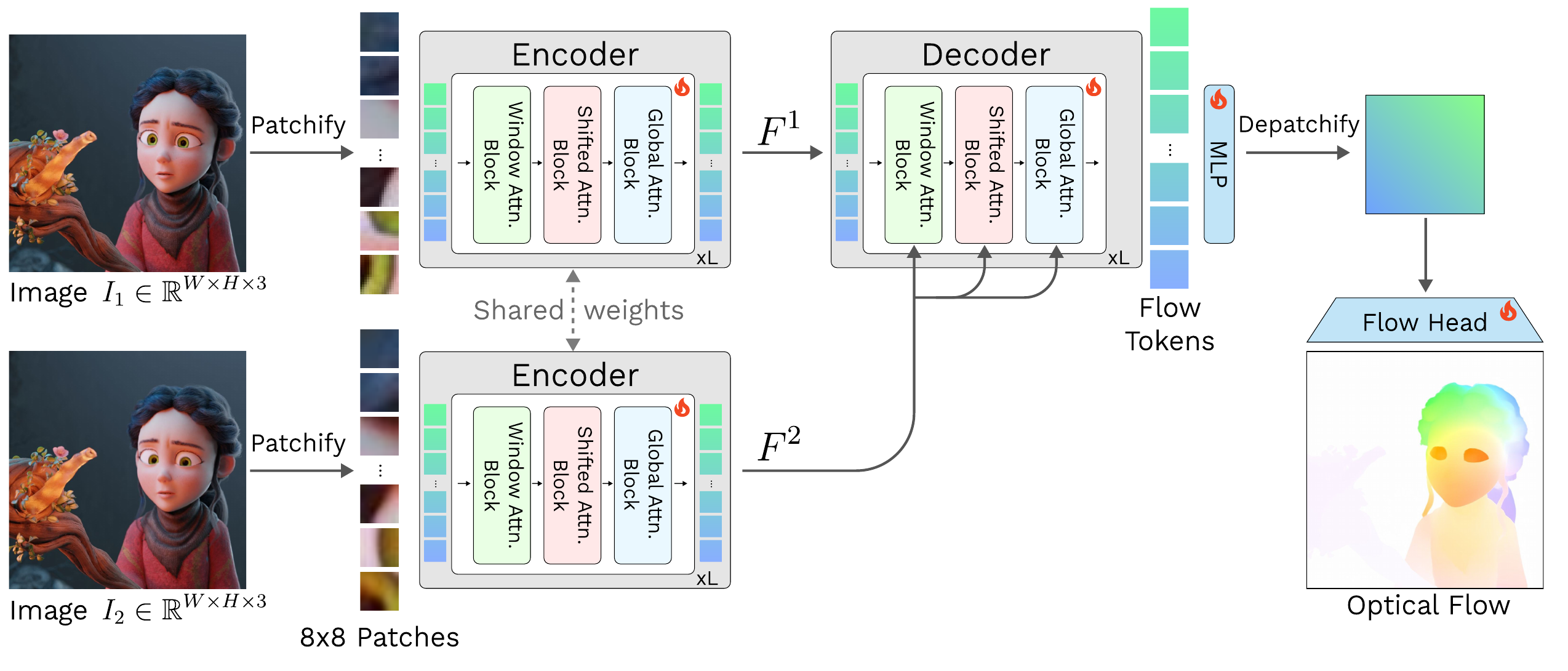}\\
\caption{\textbf{Method overview.} Given an image pair $(I_1, I_2)$, we patchify each input into $8{\times}8$ tokens and extract features using two shared-weight encoders composed of Window, Shifted-Window, and Global attention blocks (\cref{fig:attn-blocks}), producing $F^1$ and $F^2$. A transformer decoder combines self-attention with cross-attention between $F^1$ and $F^2$ to form flow tokens, which are depatchified and mapped to a dense optical flow field by a lightweight prediction head.
}
\label{fig:method}
\end{figure*}

In this section, we present FreeFlow, an inductive-bias-free transformer for optical flow estimation (\cref{fig:method}). We design this approach with two goals in mind. First, we aim to demonstrate that state-of-the-art optical flow can be achieved without the flow-specific architectural modules that dominate modern pipelines, such as correlation volumes, feature warping, iterative refinement, etc. (summarized in \cref{tab:ind-bias}). Second, we seek a design that remains practical at high resolutions and ensures repeated information exchange across processing scales.

To this end, we study existing high-resolution prediction strategies, identify their limitations, and derive a principled alternative (\cref{fig:tiling-comp}). A straightforward solution is inference-time tiling in CroCo-/FlowFormer-style pipelines (\cref{fig:tiling-comp}a), but tiles interact only through late-stage averaging, so information does not flow across tile borders during feature extraction, and the number of forward passes grows with resolution. An alternative is multi-scale decoding with \emph{Late Feature Fusion} as in DepthPro-like designs (\cref{fig:tiling-comp}b), which reduces the number of forward passes, but keeps information exchange delayed and tied to fixed-resolution assumptions, and is unreliable in the binocular setting when objects cross tile boundaries. FreeFlow resolves these issues by using a fixed and efficient tiling scheme while enabling dense feature exchange throughout the network (\cref{fig:tiling-comp}c), combining local processing with cross-tile and global interactions.

We next give a functional description of the architecture and then formalize the attention variants used by FreeFlow.

\subsection{Approach}
We adopt the CroCo/DUSt3R/MASt3R encoder--decoder high-level architecture for binocular reasoning, i.e., a Siamese ViT encoder followed by a decoder that alternates self- and cross-attention between the two views.

Given two input images $I_1, I_2 \in \mathbb{R}^{H \times W \times 3}$, we embed each image into a sequence of non-overlapping $P{\times}P$ patches (with $P{=}8$) using a standard patch projection, producing token sequences $X_1, X_2 \in \mathbb{R}^{N \times D}$ where $N=\frac{H}{P}\cdot\frac{W}{P}$. Both sequences are processed by a Siamese transformer encoder with shared weights to obtain feature representations
\begin{equation}
F^1 = \operatorname{Encoder}(X_1), \qquad F^2 = \operatorname{Encoder}(X_2),
\end{equation}
with $F^1, F^2 \in \mathbb{R}^{N \times D}$.

A transformer decoder then produces flow tokens
\begin{equation}
Z = \operatorname{Decoder}(F^1, F^2),
\end{equation}
where each decoder block combines self-attention over the current tokens with cross-attention from $F^1$ (queries) to $F^2$ (keys/values), enabling repeated information exchange between the two views. The output $Z \in \mathbb{R}^{N \times D}$ is reshaped into a spatial feature map $Z_{\mathrm{map}} \in \mathbb{R}^{\frac{H}{P}\times\frac{W}{P}\times D}$ and mapped to a dense flow and confidence field $U \in \mathbb{R}^{H \times W \times 5}$ by a prediction head.

\begin{figure*}[t]
\centering
\includegraphics[width=.9\textwidth]{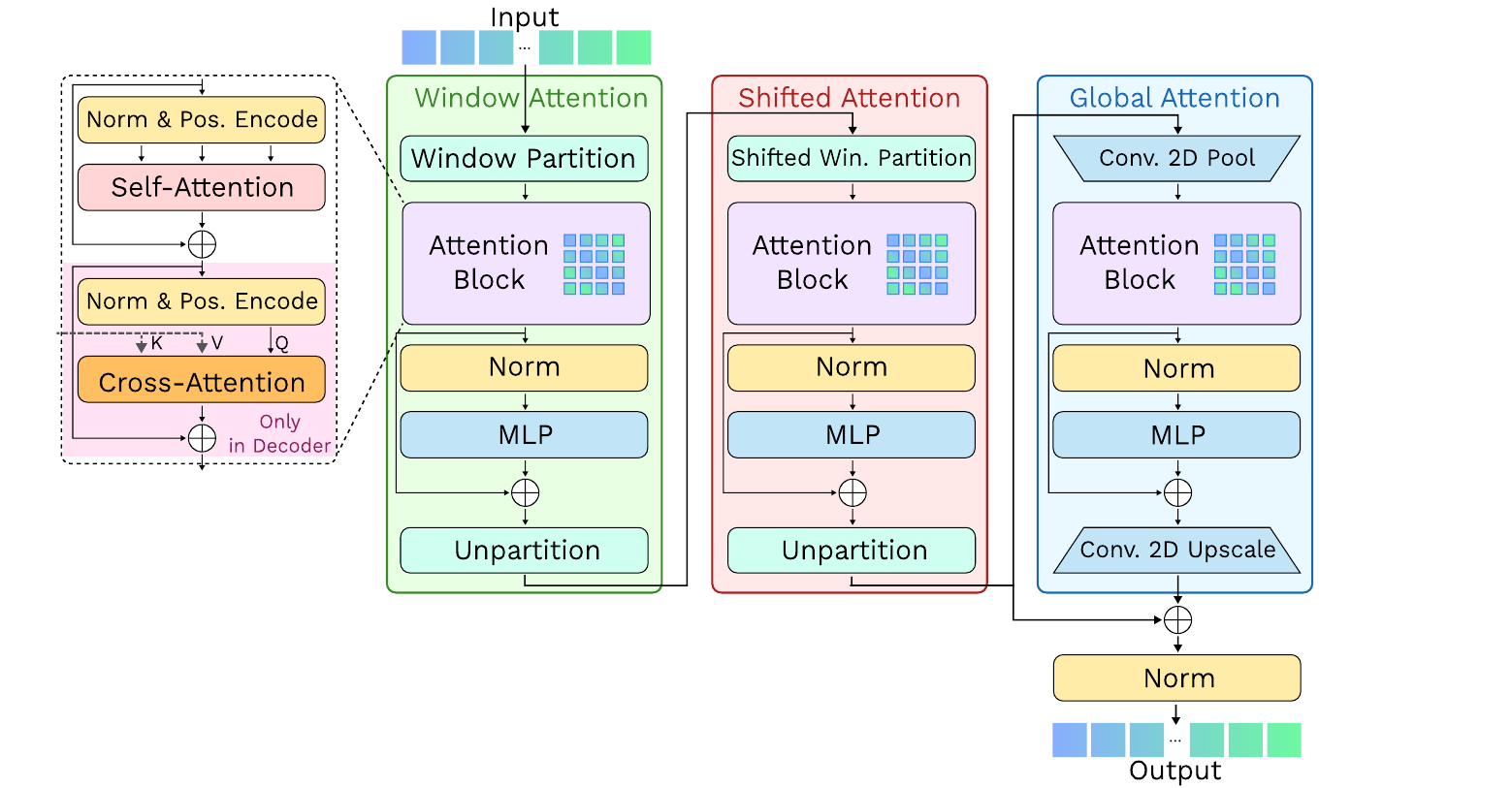}
\caption{\textbf{Attention block variants.} Our architecture uses three attention patterns: (i) \emph{Window} attention over a $4{\times}4$ partition (16 non-overlapping windows), (ii) \emph{Shifted-Window} attention with a half-window offset to exchange information across window boundaries, and (iii) \emph{Global} attention applied at $2{\times}$ lower spatial resolution via down/up-sampling. See \cref{fig:tiling-comp} for the partitioning visualization.}
\label{fig:attn-blocks}
\end{figure*}

\subsubsection{Attention Block Types.}
To predict highly-detailed globally consistent flow fields, FreeFlow uses a hierarchical attention design that mixes local processing, cross-window exchange, and global context. Each block follows the CroCo-style transformer structure: encoder blocks apply self-attention, while decoder blocks additionally use cross-attention to utilize tokens from the other view. We use three attention variants (\cref{fig:attn-blocks}):

\begin{itemize}
    \item \textit{Window (Win) Attention Block.}
    The token map of size $\frac{H}{8}\times\frac{W}{8}$ is partitioned into a $4\times4$ grid of $16$ non-overlapping windows, each containing $\frac{H}{32}\times\frac{W}{32}$ tokens. Attention is computed independently within each window.

    \item \textit{Shifted-Window (Swin) Attention Block.}
    To enable information flow across window (and tile) boundaries, we apply a half-window shift by $\left\lfloor\frac{W}{64}\right\rfloor$ tokens horizontally and $\left\lfloor\frac{H}{64}\right\rfloor$ tokens vertically, partition into the same $4\times4$ windows, perform window attention, and shift back.

    \item \textit{Global Attention Block.}
    To incorporate global context at controlled cost, we downsample by $2\times$ with a stride-2 convolution, apply full attention at the reduced resolution, and upsample by $2\times$ with a stride-2 transposed convolution. We apply normalization after the residual connection.
\end{itemize}

Each encoder/decoder layer applies the blocks in the fixed order: Win, Swin, and Global. Since attention cost scales quadratically with the number of tokens, the $2\times$ downsampling in the Global block keeps its attention cost comparable to Win/Swin at the original resolution. To encode token positional information within the image, we rely on Rotary Positional Embedding (RoPE)~\cite{rope2024}.

\subsubsection{Attention Scale Factor.} 
Following prior work\cite{dong2024memflow} on resolution-adaptive attention scaling, we multiply the attention logits by a logarithmic factor of the token count, which improves generalization when running inference at resolutions higher than those seen during training. For a token map of size $\frac{H}{8}\times\frac{W}{8}$ we use
\begin{align}
\operatorname{Self-Attention}(Q, K, V) &= \operatorname{Softmax} \left( \frac{\log(H/8 \times W/8)}{\sqrt {D}} \times Q K^T \right) \times V \\
\operatorname{Cross-Attention}(Q_1, K_2, V_2) &= \operatorname{Softmax} \left( \frac{\log(H/8 \times W/8)}{\sqrt {D}} \times Q_1 K_2^T \right) \times V_2
\end{align}

Unlike prior formulations that normalize the factor to be $1$ at the training token count, we use the unnormalized variant and found it to work well in practice.

\begin{figure*}[t]
\centering
\includegraphics[width=\textwidth]{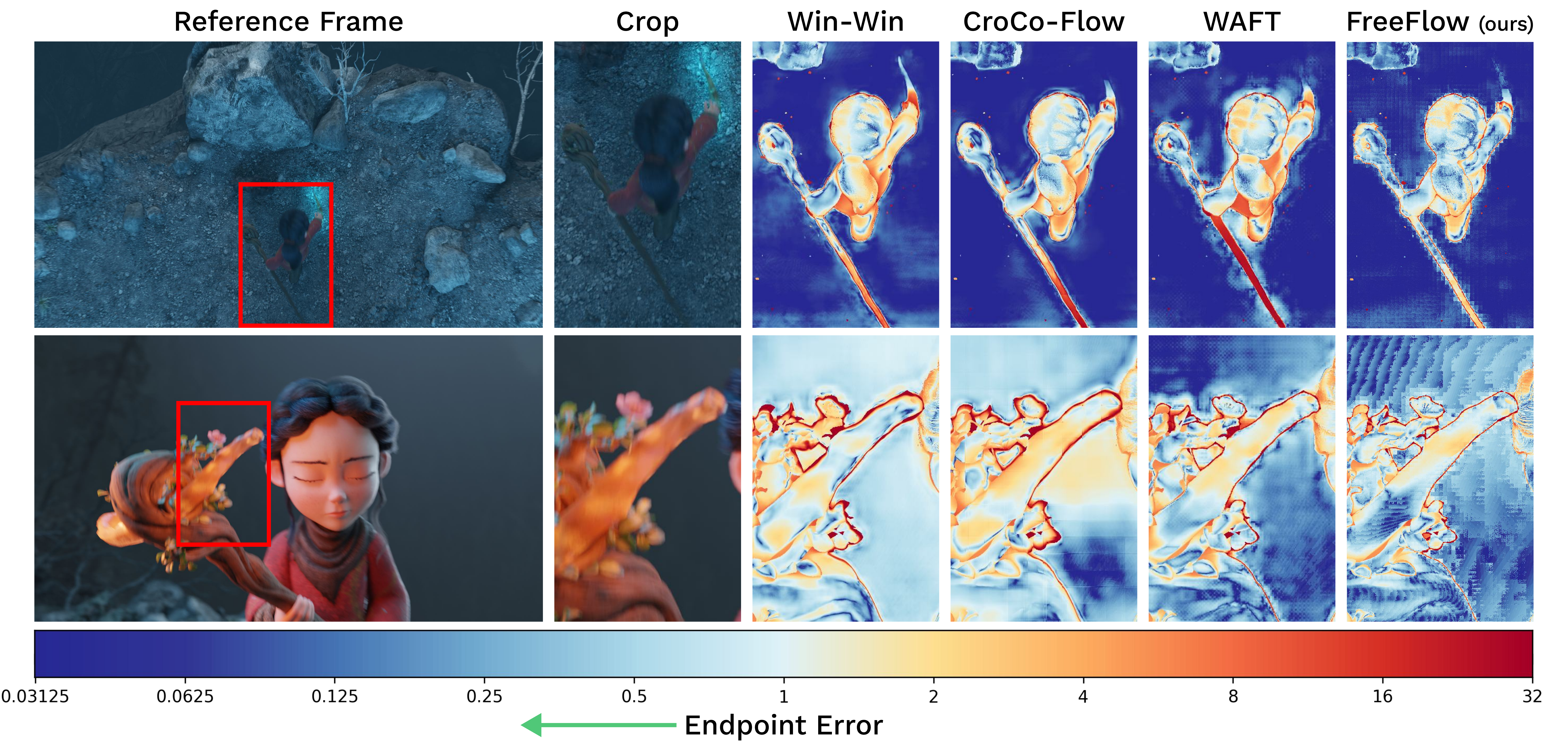}\\
\caption{\textbf{Qualitative comparison on the Spring benchmark~\cite{mehl2023spring}.} Examples of error-maps of Win-Win~\cite{leroy2023win}, CroCo-Flow~\cite{weinzaepfel2023croco}, WAFT~\cite{wang2025waft}, and FreeFlow predictions;  
the colorbar represents endpoint error. Our approach combines high level of detail (note the hole in the staff that only our method captures and the thin object bounds) with high motion consistency (note the staff's end in the top example and the low error in the background of the bottom example). Crops are sourced from official leaderboard submissions.}
\label{fig:spring_viz}
\end{figure*}

\subsubsection{Flow Head.}
Most high-performing optical flow pipelines rely on flow-specific prediction machinery, such as iterative update stages and convex upsampling. In addition, bias-free transformer baselines often use DPT-style heads from~\cite{ranftl2021vision} that aggregate features from multiple layers (and, in CroCo-style designs, may also reuse encoder features) to form the final prediction. In contrast, FreeFlow predicts flow directly from the final decoded patch map using a simple three-layer head, without iterative refinement or convex upsampling.

Given $F \in \mathbb{R}^{\frac{H}{8} \times \frac{W}{8} \times D}$, we apply a $3{\times}3$ convolution to expand channels to $4D$, followed by a $1{\times}1$ convolution to $4096$ channels, and a transposed convolution with kernel and stride $8$ to upsample to $H \times W$. The head outputs $\mathbb{R}^{H \times W \times 5}$, where the first two channels represent optical flow and the remaining three parameterize the uncertainty terms used by the mixture-of-Laplace loss (following SEA-RAFT~\cite{wang2025sea}). Finally, we multiply the flow channels by $8$ (the patch size) to obtain flow in pixel units.
\section{Experiments}
We first detail our training pipeline, consisting of cross-view completion pretraining followed by finetuning for optical flow. We evaluate our method on three popular optical flow benchmarks: Spring~\cite{mehl2023spring} (high-resolution real-world sequences), Sintel~\cite{butler2012naturalistic} (synthetic scenes with complex motion and rendering effects), and KITTI-2015~\cite{Menze2015CVPR} (real driving scenes). Finally, we provide ablations of key design choices, including the attention configuration, pretraining masking ratio, and model scaling.

\begin{table}[t]
    \caption{
    \textbf{Training procedure details.} 
    Dataset abbreviations: TA: TartanAir~\cite{wang2020tartanair}, T: Things~\cite{mayer2016large}, S: Sintel~\cite{butler2012naturalistic}, K: KITTI-2015~\cite{Menze2015CVPR}, H: HD1K~\cite{hd1k2016}. Inspired by SEA-RAFT~\cite{wang2025sea} and MEMFOF~\cite{Bargatin_2025_ICCV}, the dataset distributions for the TaTSKH stages are TA (0.23), S(0.25), T(0.24), K(0.09), H(0.19).}
    \label{tab:training}
    \centering
    \begin{tabular}{ccccccccc}
\toprule
Stage & Weights & Datasets & Scale & Crop size & LR & WD & Batch & Steps \\

\midrule
\multirow{3}{*}{Pretrain} & \multirow{3}{*}{---} & ARKitScenes~\cite{arkit2021}& \multirow{3}{*}{1x} & \multirow{3}{*}{[224, 224]} & \multirow{3}{*}{8e-4} & \multirow{3}{*}{5e-2} & \multirow{3}{*}{2048} & \multirow{3}{*}{346k} \\
& & MegaDepth~\cite{megadepth2018} & & & & & & \\
& & 3DStreetView~\cite{3dstreetview2016} & & & & & & \\
\cdashlinelr{1-9}
TaTSKH & Pretrain & TA+T+S+K+H & 2x & 10880 tok. & 4e-5 & 1e-2 & 32 & 450k \\
TaTSKH-hq & TaTSKH & TA+T+S+K+H & 2x & 32640 tok. & 1e-5 & 1e-5 & 32 & 90k \\
Sintel-ft & TaTSKH-hq & S & 2x & [872, 2048] & 1e-5 & 1e-5 & 32 & 12.5k\\
KITTI-ft & TaTSKH-hq & K & 2x & [750, 2484] & 1e-5 & 1e-5 & 32 & 2.5k \\
Spring-ft & TaTSKH-hq & Spring~\cite{mehl2023spring} & 1x & [1080, 1920] & 1e-5 & 1e-5 & 32 & 60k \\

\bottomrule
\end{tabular}
\end{table}

\subsection{Training Details}

Following CroCo~\cite{weinzaepfel2022croco, weinzaepfel2023croco}, we first pre-train our model on the cross-view completion task and then finetune the resulting weights with a new head for optical flow estimation. In cross-view completion, a large fraction of patches in one view is replaced by a learned token $e_{\text{mask}}$, and the model reconstructs the missing content conditioned on the second view. Due to its two-image nature, this objective encourages learning dense long-range correspondences, which is well aligned with the downstream task of binocular matching. Training details and datasets are summarized in \cref{tab:training}; please refer to the supplementary for additional details.

\subsubsection{Pre-train.} We follow the pre-training stage protocol from CroCo with minor adjustments. During cross-view completion training, a model takes as input two images that represent different views of the same scene: one view is severely masked, and the model has to predict the masked regions using the information from the second view. We sample image pairs from ARKitScenes~\cite{arkit2021}, MegaDepth~\cite{megadepth2018}, and 3DStreetView~\cite{3dstreetview2016}, resulting in 3.7M data samples in total. We use fixed-size crops of 224$\times$224 resolution and pretrain the model for 346k steps.

The only significant difference from the CroCo setup is when the learned masked patch representation $e_{\text{mask}}$ is introduced. In CroCo, masked tokens are removed from the first view and the corresponding $e_{\text{mask}}$ tokens are added only at the decoder input. Here, due to the hierarchical nature of our model, we replace masked patches with $e_{\text{mask}}$ at the encoder input, while keeping the completion objective unchanged.

\subsubsection{Optical Flow Finetune.}
The finetuning stage protocol is inspired by MEMFOF~\cite{Bargatin_2025_ICCV}; specifically, we adopt their $2\times$ upsampling of training frames, which better matches the motion distribution of FullHD inputs and improves high-resolution performance. Unlike MEMFOF and other curriculum-based training recipes that use multiple sequential stages, we use a single main dataset mixture, denoted TaTSKH in \cref{tab:training}, for simplicity. For additional speed, we split this finetuning into a low- and high-token-count stage (TaTSKH and TaTSKH-hq in \cref{tab:training}).

Instead of using a fixed crop size, to avoid unnecessary padding and to expose the model to a wider motion range, we use variable-resolution training with a fixed token budget per minibatch. More specifically, for each sample we randomly choose one spatial dimension (height or width), sample its value, and set the other dimension to the largest value such that the resulting token count does not exceed the prescribed budget. This produces rectangular crops with varying aspect ratios while keeping compute and memory controlled. The implementation is straightforward as we use a batch size of 1 sample/GPU during the finetuning stage. For benchmark submissions, we further finetune with fixed crop sizes (Sintel-ft, KITTI-ft, Spring-ft in \cref{tab:training}). Following SEA-RAFT\cite{wang2025sea}, we use the Mixture-of-Laplace loss.

In total, it takes from 4 to 5 days to pre-train and around 3 days to finetune our largest model on 32 GPUs.

\subsection{Results}
We adopt four widely used metrics from established benchmarks in this study: endpoint error (EPE), 1-pixel outlier rate (1px), Fl-score, and WAUC error. Please refer to \cite{richter2017playing, butler2012naturalistic, mehl2023spring, morimitsu2025dpflow, Menze2015CVPR} or the supplementary for their definitions.

\begin{table*}[t!]
    \caption{\textbf{Spring benchmark results.} 
    "*" indicates that it was submitted by the Spring team without being finetuned on the provided training set. Speed (runtime) and peak GPU memory consumption were measured on a Nvidia RTX 3090 GPU (24 GB) with automatic mixed precision and without memory efficient correlation volumes, where "---" denotes no available data or implementation for a given model and "\textdagger" denotes our best estimate based on the authors description. "MF" indicates that the method uses multiple frames (three or more). The best results are indicated in \textbf{bold}, second-best are \underline{underlined}, third best are indicated in \textit{italic}. Method configurations are taken from submissions to the Spring benchmark if present, and from submissions to the Sintel benchmark otherwise.}
    \label{tab:spring-comparison}
    \centering
    \resizebox{1.0\linewidth}{!}{
\begin{tabular}{l c c c c c c c}
\toprule
\multirow{2}[2]{*}{Method} & \multicolumn{2}{c}{Inf. Cost (1080p)} & \multirow{2}[2]{*}{Params (M)} & \multicolumn{4}{c}{Spring} \\
\cmidrule(lr){2-3}\cmidrule(lr){5-8}
 & Memory (GB) & Time (ms) & & 1px $\downarrow$ & EPE $\downarrow$ & Fl $\downarrow$ & WAUC $\uparrow$ \\
\midrule
FlowNet2~\cite{ilg2017flownet2} & 3.01 & 110 & 162.52 & 6.710\rlap{$^{*}$} & 1.040\rlap{$^{*}$} & 2.823\rlap{$^{*}$} & 90.907\rlap{$^{*}$} \\
PWC-Net~\cite{sun2018pwc} & 0.57 & 48 & 9.37 & 82.265\rlap{$^{*}$} & 2.288\rlap{$^{*}$} & 4.889\rlap{$^{*}$} & 45.670\rlap{$^{*}$} \\
RAFT~\cite{teed2020raft} & 7.97 & 406 & 5.26 & 6.790\rlap{$^{*}$} & 1.476\rlap{$^{*}$} & 3.198\rlap{$^{*}$} & 90.920\rlap{$^{*}$} \\
GMA~\cite{jiang2021learning_GMA} & 11.81 & 830 & 5.88 & 7.074\rlap{$^{*}$} & 0.914\rlap{$^{*}$} & 3.079\rlap{$^{*}$} & 90.722\rlap{$^{*}$} \\
GMFlow~\cite{xu2022gmflow} & 8.22 & 8024 & 4.72 & 10.355\rlap{$^{*}$} & 0.945\rlap{$^{*}$} & 2.952\rlap{$^{*}$} & 82.337\rlap{$^{*}$} \\
FlowFormer~\cite{huang2022flowformer} & 1.90 & 2084\rlap{\textsuperscript{\textdagger}} & 16.17 & 6.510\rlap{$^{*}$} & 0.723\rlap{$^{*}$} & 2.384\rlap{$^{*}$} & 91.679\rlap{$^{*}$} \\
SEA-RAFT~(M)~\cite{wang2025sea} & 8.12 & 198 & 19.67 & 3.686 & 0.363 & 1.347 & 94.534 \\
DPFlow~\cite{morimitsu2025dpflow}\ & 4.26 & 401 & 10.02 & 3.442 & 0.340 & 1.311 & 94.980 \\
MemFlow\textsuperscript{(MF)}\cite{dong2024memflow} & 8.06 & 754 & 6.27 & 4.482 & 0.471 & 1.416 & 93.855 \\
StreamFlow\textsuperscript{(MF)}\cite{sun2025streamflow} & 18.61 & 898 & 14.25 & 4.152 & 0.467 & 1.424 & 94.404 \\
MEMFOF\textsuperscript{(MF)}\cite{Bargatin_2025_ICCV} & 1.90 & 262 & 75.78 & 3.289 & 0.355 & 1.238 & \textit{95.186} \\
ARFlow\textsuperscript{(MF)}\cite{liuarflow} & --- & --- & 76.50 & \textit{3.265} & 0.353 & 1.212 & \textbf{95.283} \\
CroCo-Flow~\cite{weinzaepfel2023croco} & 2.73 & 3266 & 447.47 & 4.565 & 0.498 & 1.508 & 93.660 \\
Win-Win~\cite{leroy2023win} & 3.82\rlap{\textsuperscript{\textdagger}} & 305\rlap{\textsuperscript{\textdagger}} &
229.77\rlap{\textsuperscript{\textdagger}} & 5.371 & 0.475 & 1.621 & 92.720 \\
WAFT-DAv2-a2~\cite{wang2025waft} & 20.58 & 489 & 56.93 & 3.298 & \underline{0.304} & \textit{1.197} & 94.990 \\
WAFT-DINOv3-a2~\cite{wang2025waft} & 18.96 & 408 & 56.47 & \textbf{3.182} & \textit{0.325} & 1.246 & 95.051 \\
FreeFlow-S (ours) & 1.02 & 144 & 34.58 & 5.087 & 0.533 & 1.452 & 90.196 \\
FreeFlow-M (ours) & 1.66 & 325 & 102.46 & 3.392 & 0.346 & \underline{1.171} & 94.919 \\
FreeFlow-L (ours) & 2.58 & 607 & 230.72 & \underline{3.192} & \textbf{0.278} & \textbf{1.048} & \underline{95.235} \\
\bottomrule
\end{tabular}
}
\end{table*}

\subsubsection{Results on Spring.} FreeFlow achieves state-of-the-art performance on Spring. FreeFlow-L sets the best EPE and Fl among all compared methods (Tab.~\ref{tab:spring-comparison}), while remaining on par with the strongest approaches in 1px and WAUC; in particular, it attains the best WAUC among two-frame methods. Compared to WAFT-DAv2-a2, FreeFlow-L improves EPE by 9\% and reduces Fl by 14\%. Owing to tiling-free native 1080p inference, FreeFlow preserves fine detail while maintaining global motion consistency (\cref{fig:spring_viz}). We further highlight that native 1080p processing is possible within a low inference memory budget (\cref{fig:intro}).

\begin{table}[t]
    \begin{minipage}{0.54\textwidth}
    \caption{\textbf{Sintel~\cite{butler2012naturalistic} and KITTI-15~\cite{Menze2015CVPR} benchmark results.} 
    Sintel uses EPE as it's metric for both splits, while KITTI-15 uses the Fl-all outliers metric. "---" indicates no published results and "MF" indicates that the method used multiple frames (three or more) to generate it's submissions.}
    \label{tab:sintel-kitti}
    \centering
    \resizebox{1.0\linewidth}{!}{
\begin{tabular}{l c c c}
\toprule
\multirow{2}[2]{*}{Method} & \multicolumn{2}{c}{Sintel} & \multicolumn{1}{c}{KITTI-15}\\
\cmidrule(lr){2-3} \cmidrule(lr){4-4}
 & Clean $\downarrow$ & Final $\downarrow$ & Fl-all $\downarrow$ \\
\midrule
FlowNet2~\cite{ilg2017flownet2} & 4.16 & 5.74 & 10.41 \\
PWC-Net~\cite{sun2018pwc} & 3.90 & 5.04 & 9.60 \\
RAFT~\cite{teed2020raft} & 1.61 & 2.86 & 5.10 \\
GMA~\cite{jiang2021learning_GMA} & 1.39 & 2.47 & 5.15 \\
GMFlow+~\cite{xu2023unifying} & 1.03 & 2.37 & 4.49 \\
FlowFormer~\cite{huang2022flowformer} & 1.16 & 2.09 & 4.68 \\  
FlowFormer++~\cite{shi2023flowformer++} & 1.07 & 1.94 & 4.52 \\
TransFlow~\cite{lu2023transflow} & 1.06 & 2.08 & 4.32 \\ 
SEA-RAFT~(L)~\cite{wang2025sea} & 1.31 & 2.60 & 4.30 \\
DPFlow~\cite{morimitsu2025dpflow} & 1.05 & 1.98 & 3.56 \\
VideoFlow-BOF\textsuperscript{(MF)}\cite{shi2023videoflow} & 1.00 & \textit{1.71} & 4.44 \\
VideoFlow-MOF\textsuperscript{(MF)}\cite{shi2023videoflow} & 0.99 & \underline{1.65} & 3.65 \\
MEMFOF\textsuperscript{(MF)}\cite{Bargatin_2025_ICCV} & 0.99 & 1.94 & \underline{2.94} \\
MEMFOF-XL\textsuperscript{(MF)}\cite{Bargatin_2025_ICCV} & 0.93 & 1.89 & --- \\
ARFlow\textsuperscript{(MF)}\cite{liuarflow} & 0.96 & 1.79 & \textbf{2.85} \\
DDVM\cite{saxena2023surprising} & 1.75 & 2.48 & 3.26 \\
CroCo-Flow\cite{weinzaepfel2023croco} & 1.09 & 2.44 & 3.64 \\
Win-Win\cite{leroy2023win} & 1.15 & 2.34 & --- \\
WAFT-DAv2-a2\cite{wang2025waft} & 0.94 & 2.33 & 3.31 \\
WAFT-DINOv3-a2\cite{wang2025waft} & 0.95 & 2.02 & 3.56 \\
GeoVIT \cite{wu2025study} & \underline{0.79} & 1.88 & 3.79 \\
FreeFlow-S (ours) & 1.03 & 1.99 & 4.06 \\
FreeFlow-M (ours) & \textit{0.80} & 1.77 & 3.33 \\
FreeFlow-L (ours) & \textbf{0.68} & \textbf{1.48} & \textit{3.23} \\
\bottomrule
\end{tabular}
}
    \end{minipage}
    \hfill
    \begin{minipage}{0.44\textwidth}
    \centering
    \includegraphics[width=0.95\linewidth]{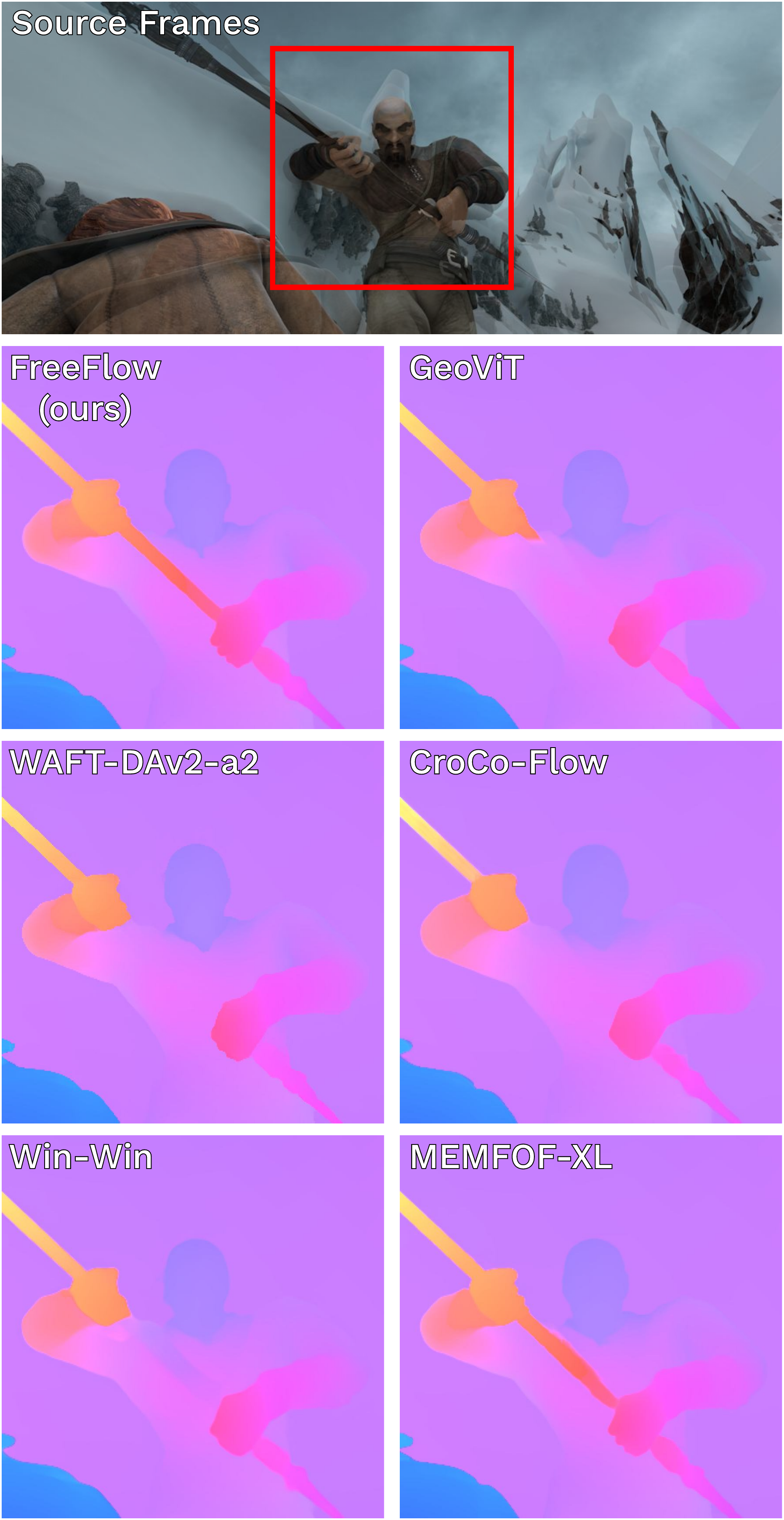}
    \captionof{figure}{\textbf{Qualitative comparison on Sintel.} Examples of FreeFlow-L, GeoViT~\cite{wu2025study}, WAFT-DAv2-a2~\cite{wang2025waft}, CroCo-Flow~\cite{weinzaepfel2023croco}, Win-Win~\cite{leroy2023win}, and MEMFOF-XL~\cite{Bargatin_2025_ICCV} outputs on the Sintel~\cite{butler2012naturalistic} benchmark. Sourced from the official leaderboard submissions.}
    \label{fig:sintel-qual}
    \end{minipage}
\end{table}

\subsubsection{Results on Sintel and KITTI.}
Following MEMFOF, we finetune on $2\times$ upsampled frames. Accordingly, for Sintel and KITTI submissions we upscale input images by $2\times$ and downscale the predicted flow by $2\times$. FreeFlow-L ranks first on Sintel on both Clean and Final (Tab.~\ref{tab:sintel-kitti}), improving over GeoVIT~\cite{wu2025study} by 14\% on Clean (0.79$\rightarrow$0.68) and 10\% over VideoFlow-MOF~\cite{shi2023videoflow} on Final (1.65$\rightarrow$1.48). Notably, FreeFlow-M is already highly competitive: it is second only to FreeFlow-L on Clean, and ranks fourth on Final, surpassed only by the 3- and 5-frame VideoFlow variants. On KITTI-2015, FreeFlow-L achieves 3.23 Fl-all, outperforming all non-stereo and non-multiframe methods on KITTI-15 (Tab.~\ref{tab:sintel-kitti}). Qualitative results show that the model captures complex motion patterns using only two frames (\cref{fig:sintel-qual}). Additional visual comparisons and zero-shot evaluations are provided in the supplementary material.

\begin{table}[t]
    \begin{minipage}{0.49\textwidth}
    \caption{
    \textbf{Masking ratio and attention-block ablation.} Spring sub-validation results for FreeFlow variants with different cross-view completion masking ratios and active attention subblocks (Win/Swin/Global). \colorbox{lightgray}{Gray} marks the configuration selected for the main model; best and second-best results are shown in \textbf{bold} and \underline{underlined}, respectively.
    }
    \label{tab:abl:arch-ablation}
    \centering
\begin{tabular}{l c c c c c}
\toprule
\multirow{2}[1]{*}{\makecell[l]{Mask.\\Ratio}} & \multicolumn{3}{c}{Subblock type} & \multicolumn{2}{c}{Spring (sub-val)}\\
\cmidrule(lr){2-4}  \cmidrule(lr){5-6}
 & Win & Swin & Global & 1px $\downarrow$ & EPE $\downarrow$ \\
\midrule
0.9 & $\times$ & $\times$ & $\checkmark$ & 1.133 & 0.229  \\
0.9 & $\checkmark$ & $\times$ & $\checkmark$ & 0.801 & 0.190 \\
0.9 & $\checkmark$ & $\checkmark$ & $\times$ & 0.659 & 0.167 \\
0.9 & $\checkmark$ & $\checkmark$ & $\checkmark$ & 0.688 & 0.170 \\
0.925 & $\checkmark$ & $\checkmark$ & $\checkmark$ & 0.685 & \underline{0.166} \\
0.95 & $\checkmark$ & $\checkmark$ & $\times$ & 0.658 & \underline{0.166} \\
\rowcolor{lightgray}
0.95 & $\checkmark$ & $\checkmark$ & $\checkmark$ & \textbf{0.624} & \textbf{0.157} \\
0.975 & $\checkmark$ & $\checkmark$ & $\checkmark$ & \underline{0.656}  & 0.174 \\
\bottomrule
\end{tabular}
    \end{minipage}
    \hfill
    \begin{minipage}{0.49\textwidth}
    \centering
    \includegraphics[width=0.9\linewidth]{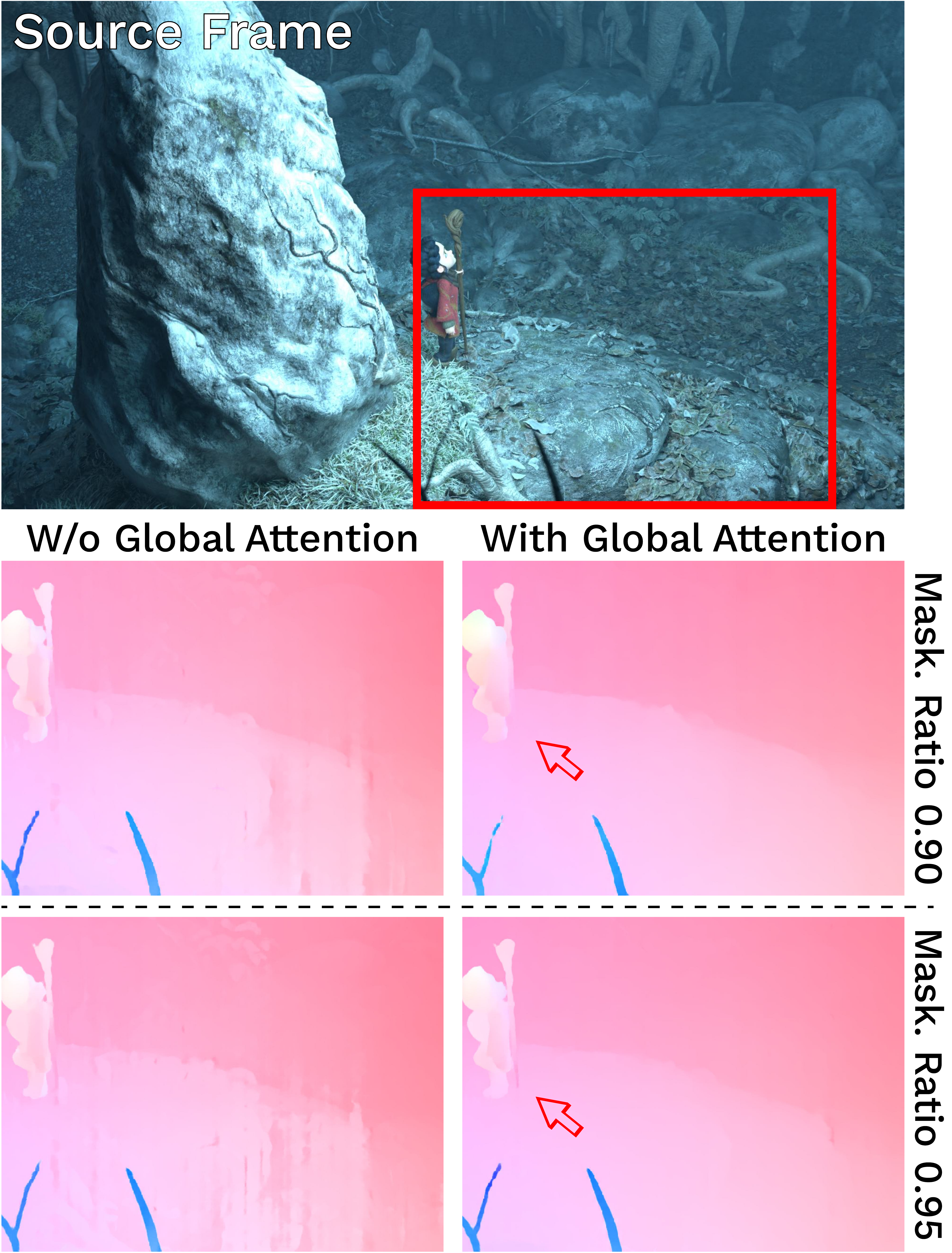}
    \captionof{figure}{
    \textbf{Qualitative ablation} comparison of FreeFlow models with and without Global Attention block and pretrained with different masking ratios on the Spring benchmark. Zoom in for better view.}
    
    \label{fig:abl:abl_examples}
    \end{minipage}
\end{table}

\subsection{Ablation Study}
Unless stated otherwise, all ablations use a scaled-down FreeFlow configuration with 8 encoder and 8 decoder layers, width 256, and 8 attention heads, and are trained with the same training recipe. Following SEA-RAFT and WAFT, we report results on the Spring sub-validation split (scenes 0045 and 0047) after finetuning on the remaining training data. Additional experiments in the supplementary isolate the architecture from the training procedure and study the effect of adding iterative flow-specific biases back into FreeFlow.

\subsubsection{Architecture and Masking Ratio Ablation.} 
We study the interaction between the cross-view completion masking ratio and the hierarchical attention design of FreeFlow (Tab.~\ref{tab:abl:arch-ablation}). The masking ratio controls the fraction of patches in the target view that are replaced by the learned token $e_{\text{mask}}$ during pretraining, while the attention configuration determines which subblock types (Win/Swin/Global) are present in the encoder and decoder. 

With all three subblocks enabled, a masking ratio of 0.95 performs best, improving over the CroCo default 0.9 by 9.3\% in 1px and 7.6\% in EPE. This is consistent with FreeFlow using smaller patches than CroCo ($8{\times}8$ vs.\ $16{\times}16$), which reduces the distance to visible regions and makes a higher mask rate beneficial as it makes the pretraining task sufficiently challenging.

We also observe an interaction between masking ratio and attention configuration. At the CroCo default ratio (0.9), removing Global does not hurt and can even slightly improve the sub-validation metrics, whereas at 0.95 Global becomes important for the best performance. This indicates that the masking ratio is an important hyperparameter that should be chosen jointly with the model architecture. Shifted-Window attention consistently contributes to accuracy, and removing it leads to a clear drop on this split. Additionally, for both masking ratios, qualitative comparisons (Fig.~\ref{fig:abl:abl_examples}) show that removing the Global block can introduce obvious motion inconsistencies, even when the sub-validation metrics change only marginally.

\begin{table}[t]
    \begin{minipage}{0.54\textwidth}
    \caption{\textbf{Model configurations.} Architectural hyperparameters of FreeFlow variants and our \emph{best-effort  estimates} for transformer-based baselines. ``+'' denotes separate encoder and decoder widths (or counts), while ``$\times$'' denotes repeated recurrent decoder calls.}
    \label{tab:abl:scale-comparison}
    \centering
    \resizebox{1.0\linewidth}{!}{
\begin{tabular}{l c c c c c c}
\toprule
\multirow{2}{*}{\rotatebox{0}{Name}} &
\multirow{2}{*}{\rotatebox{0}{\makecell[c]{Patch\\Sizes}}} &
\multirow{2}{*}{\rotatebox{0}{Width}} &
\multirow{2}{*}{\rotatebox{0}{\makecell[c]{Attn.\\Count}}} &
\multirow{2}{*}{\rotatebox{0}{\makecell[c]{Attn.\\Heads}}} &
\multirow{2}{*}{\rotatebox{0}{\makecell[c]{MLP\\Count}}} &
\multirow{2}{*}{\rotatebox{0}{\makecell[c]{Params\\(M)}}} \\
 & & & & &\\
\midrule
GeoViT & 16 & 1024 & 24$\times$6 & 16 & 24$\times$6 & 377 \\
WAFT & 16 & 384$+$384 & 12$+$12$\times$5 & 6$+$6 & 12+12$\times$5 & 56 \\
CroCo-Flow & 16 & 1024$+$768 & 24$+$24 & 16$+$12 & 24$+$12 & 447 \\
Win-Win & 16 & 768$+$768 & 12$+$24 & 12 & 12$+$12 & 230 \\
GMFlow+ & 8/4 & 0$+$128 & 0$+$12$\times$2 & 0$+$1 & 0$+$6$\times$2 & 4.7 \\
\midrule
FreeFlow-S & 8 & 256$+$256 & 12$+$24 & 4+4 & 12$+$12 & 35 \\
FreeFlow-M & 8 & 384$+$384 & 18$+$36 & 6+6 & 18$+$18 & 102 \\
FreeFlow-L & 8 & 512$+$512 & 24$+$48 & 8+8 & 24$+$24 & 231 \\
\bottomrule
\end{tabular}
}
    \end{minipage}
    \hfill
    \begin{minipage}{0.44\textwidth}
    \vspace{1em}
    \centering
    \includegraphics[width=1.0\linewidth]{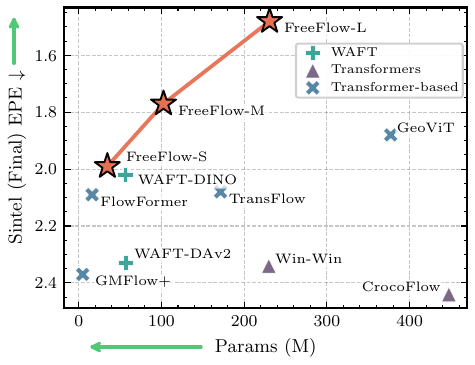}
    \captionof{figure}{\textbf{Model scaling on Sintel.} Sintel Final EPE ($\downarrow$) versus parameter count (M), $\times10^6$ for FreeFlow variants and transformer-based baselines.}
    \label{fig:scatter_ablation}
    \vspace{-1em}
    \end{minipage}
\end{table}

\subsubsection{Model Scaling.}
A benefit of FreeFlow's simple, uniform encoder--decoder design is that it can be scaled in a straightforward manner by adjusting depth and width. We therefore study how performance evolves across model sizes (S/M/L) on common optical flow benchmarks. We follow common ViT scaling~\cite{Zhai_2022_CVPR} best practices by co-scaling depth, width, and the number of attention heads while keeping the overall architecture and patch size fixed. The resulting configurations are summarized in \cref{tab:abl:scale-comparison}.

As shown in ~\cref{fig:scatter_ablation}, FreeFlow scales predictably: larger variants yield consistent accuracy gains, while smaller variants retain strong performance, indicating that the approach remains effective even when scaled down. Memory scaling with model size is reported in \cref{fig:intro:scatter}.

\section{Conclusion}
In this work, we introduced FreeFlow, a bias-free hierarchical transformer for optical flow estimation that removes conventional flow-specific design components such as correlation volumes, feature warping, and iterative refinement. Instead, FreeFlow relies on a simple feed-forward encoder--decoder architecture that combines window, shifted-window, and reduced-resolution global attention to capture motion across multiple spatial scales. This design provides a flexible and scalable framework that improves consistently with model capacity while maintaining efficient high-resolution inference.

We show that these standard optical-flow inductive biases are not required to achieve top performance. FreeFlow reaches state-of-the-art results on all popular benchmarks, including Sintel, KITTI-2015, and Spring, demonstrating that strong motion estimation can be obtained from a general-purpose transformer architecture without specialized flow modules. We hope that this work encourages further exploration of simpler and more general architectures for motion estimation and related dense correspondence tasks.

\section*{Acknowledgements}
The work of Vladislav Bargatin, Alexander Yakovenko and Khaled Abud was supported by the The Ministry of Economic Development of the Russian\linebreak Federation in accordance with the subsidy agreement (agreement identifier\linebreak 000000C313925P4H0002; grant No 139-15-2025-012). The research was carried out using the MSU-270 supercomputer of Lomonosov Moscow State University.

%
%
\bibliographystyle{splncs04}
\bibliography{main}

@String(CVPR  = {IEEE Conf. Comput. Vis. Pattern Recog.})

@String(ICCV  = {Int. Conf. Comput. Vis.})

@String(ECCV  = {Eur. Conf. Comput. Vis.})

@String(CVPRW = {IEEE Conf. Comput. Vis. Pattern Recog. Worksh.})

@String(AAAI  = {AAAI})

@String(IJCAI = {IJCAI})

@String(CVPR  = {CVPR})

@String(ICCV  = {ICCV})

@String(ECCV  = {ECCV})

@String(CVPRW = {CVPRW})

@inproceedings{chan2021basicvsr,
  author={Chan, Kelvin C.K. and Wang, Xintao and Yu, Ke and Dong, Chao and Loy, Chen Change},
  booktitle={2021 IEEE/CVF Conference on Computer Vision and Pattern Recognition (CVPR)}, 
  title={{BasicVSR}: The Search for Essential Components in Video Super-Resolution and Beyond}, 
  year={2021},
  volume={},
  number={},
  pages={4945-4954},
  xxkeywords={Computer vision;Systematics;Superresolution;Pipelines;Noise reduction;Computer architecture;Pattern recognition},
  doi={10.1109/CVPR46437.2021.00491}
}

@inproceedings{lucas1981iterative,
  TITLE = {{An Iterative Image Registration Technique with an Application to Stereo Vision}},
  AUTHOR = {Lucas, Bruce D and Kanade, Takeo},
  BOOKTITLE = {{IJCAI'81: 7th international joint conference on Artificial intelligence}},
  ADDRESS = {Vancouver, Canada},
  VOLUME = {2},
  PAGES = {674-679},
  YEAR = {1981},
  xxMONTH = Aug,
  HAL_ID = {hal-03697340},
  HAL_VERSION = {v1},
}

@article{horn1981determining,
title = {Determining optical flow},
journal = {Artificial Intelligence},
volume = {17},
number = {1-3},
pages = {185-203},
year = {1981},
issn = {0004-3702},
doi = {10.1016/0004-3702(81)90024-2},
author = {Berthold K.P. Horn and Brian G. Schunck},
xxabstract = {Optical flow cannot be computed locally, since only one independent measurement is available from the image sequence at a point, while the flow velocity has two components. A second constraint is needed. A method for finding the optical flow pattern is presented which assumes that the apparent velocity of the brightness pattern varies smoothly almost everywhere in the image. An iterative implementation is shown which successfully computes the optical flow for a number of synthetic image sequences. The algorithm is robust in that it can handle image sequences that are quantized rather coarsely in space and time. It is also insensitive to quantization of brightness levels and additive noise. Examples are included where the assumption of smoothness is violated at singular points or along lines in the image.}
}

@inproceedings{dosovitskiy2015flownet,
  author={Dosovitskiy, Alexey and Fischer, Philipp and Ilg, Eddy and Häusser, Philip and Hazirbas, Caner and Golkov, Vladimir and Smagt, Patrick van der and Cremers, Daniel and Brox, Thomas},
  booktitle={2015 IEEE International Conference on Computer Vision (ICCV)}, 
  title={{FlowNet}: Learning Optical Flow with Convolutional Networks}, 
  year={2015},
  volume={},
  number={},
  pages={2758--2766},
  xxkeywords={Optical imaging;Computer architecture;Image resolution;Correlation;Optical fiber networks;Neural networks;Optical computing},
  doi={10.1109/ICCV.2015.316}}

@inproceedings{teed2020raft,
author="Teed, Zachary
and Deng, Jia",
xxeditor="Vedaldi, Andrea
and Bischof, Horst
and Brox, Thomas
and Frahm, Jan-Michael",
title="{RAFT}: Recurrent All-Pairs Field Transforms for Optical Flow",
booktitle="Computer Vision -- ECCV 2020",
year="2020",
publisher="Springer International Publishing",
address="Cham",
pages="402--419",
isbn="978-3-030-58536-5",
doi={10.1007/978-3-030-58536-5_24}
}

@inproceedings{wang2025sea,
author="Wang, Yihan
and Lipson, Lahav
and Deng, Jia",
xxeditor="Leonardis, Ale{\v{s}}
and Ricci, Elisa
and Roth, Stefan
and Russakovsky, Olga
and Sattler, Torsten
and Varol, G{\"u}l",
title="{SEA-RAFT}: Simple, Efficient, Accurate RAFT for Optical Flow",
booktitle="Computer Vision -- ECCV 2024",
year="2025",
publisher="Springer Nature Switzerland",
address="Cham",
pages="36--54",
isbn="978-3-031-72667-5",
doi={10.1007/978-3-031-72667-5_3}
}

@article{xu2023unifying,
  author={Xu, Haofei and Zhang, Jing and Cai, Jianfei and Rezatofighi, Hamid and Yu, Fisher and Tao, Dacheng and Geiger, Andreas},
  journal={IEEE Transactions on Pattern Analysis and Machine Intelligence}, 
  title={Unifying Flow, Stereo and Depth Estimation}, 
  year={2023},
  volume={45},
  number={11},
  pages={13941-13958},
  xxkeywords={Task analysis;Optical flow;Costs;Estimation;Three-dimensional displays;Transformers;Solid modeling;Cross-attention;dense correspondence;depth;optical flow;stereo;transformer},
  doi={10.1109/TPAMI.2023.3298645}}

@inproceedings{butler2012naturalistic,
author="Butler, Daniel J.
and Wulff, Jonas
and Stanley, Garrett B.
and Black, Michael J.",
xxeditor="Fitzgibbon, Andrew
and Lazebnik, Svetlana
and Perona, Pietro
and Sato, Yoichi
and Schmid, Cordelia",
title="A Naturalistic Open Source Movie for Optical Flow Evaluation",
booktitle="Computer Vision -- ECCV 2012",
year="2012",
publisher="Springer Berlin Heidelberg",
address="Berlin, Heidelberg",
pages="611--625",
xxabstract="Ground truth optical flow is difficult to measure in real scenes with natural motion. As a result, optical flow data sets are restricted in terms of size, complexity, and diversity, making optical flow algorithms difficult to train and test on realistic data. We introduce a new optical flow data set derived from the open source 3D animated short film Sintel. This data set has important features not present in the popular Middlebury flow evaluation: long sequences, large motions, specular reflections, motion blur, defocus blur, and atmospheric effects. Because the graphics data that generated the movie is open source, we are able to render scenes under conditions of varying complexity to evaluate where existing flow algorithms fail. We evaluate several recent optical flow algorithms and find that current highly-ranked methods on the Middlebury evaluation have difficulty with this more complex data set suggesting further research on optical flow estimation is needed. To validate the use of synthetic data, we compare the image- and flow-statistics of Sintel to those of real films and videos and show that they are similar. The data set, metrics, and evaluation website are publicly available.",
isbn="978-3-642-33783-3",
doi={10.1007/978-3-642-33783-3_44}
}

@inproceedings{mehl2023spring,
  author={Mehl, Lukas and Schmalfuss, Jenny and Jahedi, Azin and Nalivayko, Yaroslava and Bruhn, Andrés},
  booktitle={2023 IEEE/CVF Conference on Computer Vision and Pattern Recognition (CVPR)}, 
  title={{Spring}: A High-Resolution High-Detail Dataset and Benchmark for Scene Flow, Optical Flow and Stereo}, 
  year={2023},
  volume={},
  number={},
  pages={4981-4991},
  xxkeywords={Computer vision;Image motion analysis;Image resolution;Estimation;Training data;Benchmark testing;Visual effects;Datasets and evaluation},
  doi={10.1109/CVPR52729.2023.00482}}

@inproceedings{ryali2023hiera,
  title = 	 {{Hiera}: A Hierarchical Vision Transformer without the Bells-and-Whistles},
  author =       {Ryali, Chaitanya and Hu, Yuan-Ting and Bolya, Daniel and Wei, Chen and Fan, Haoqi and Huang, Po-Yao and Aggarwal, Vaibhav and Chowdhury, Arkabandhu and Poursaeed, Omid and Hoffman, Judy and Malik, Jitendra and Li, Yanghao and Feichtenhofer, Christoph},
  booktitle = 	 {Proceedings of the 40th International Conference on Machine Learning},
  pages = 	 {29441--29454},
  year = 	 {2023},
  xxeditor = 	 {Krause, Andreas and Brunskill, Emma and Cho, Kyunghyun and Engelhardt, Barbara and Sabato, Sivan and Scarlett, Jonathan},
  volume = 	 {202},
  series = 	 {Proceedings of Machine Learning Research},
  xxmonth = 	 {23--29 Jul},
  publisher =    {PMLR}
}

@inproceedings{Menze2015CVPR,
  author={Menze, Moritz and Geiger, Andreas},
  booktitle={2015 IEEE Conference on Computer Vision and Pattern Recognition (CVPR)}, 
  title={Object scene flow for autonomous vehicles}, 
  year={2015},
  volume={},
  number={},
  pages={3061-3070},
  xxkeywords={Three-dimensional displays;Solid modeling;Benchmark testing;Design automation;Optical imaging;Optical sensors;Vehicle dynamics},
  doi={10.1109/CVPR.2015.7298925}
}

@inproceedings{sun2018pwc,
  author={Sun, Deqing and Yang, Xiaodong and Liu, Ming-Yu and Kautz, Jan},
  booktitle={2018 IEEE/CVF Conference on Computer Vision and Pattern Recognition}, 
  title={{PWC-Net}: CNNs for Optical Flow Using Pyramid, Warping, and Cost Volume}, 
  year={2018},
  volume={},
  number={},
  pages={8934-8943},
  xxkeywords={Optical imaging;Adaptive optics;Benchmark testing;Computational modeling;Task analysis;Feature extraction;Computer vision},
  doi={10.1109/CVPR.2018.00931}}

@inproceedings{dong2024memflow,
  author={Dong, Qiaole and Fu, Yanwei},
  booktitle={2024 IEEE/CVF Conference on Computer Vision and Pattern Recognition (CVPR)}, 
  title={{MemFlow}: Optical Flow Estimation and Prediction with Memory}, 
  year={2024},
  volume={},
  number={},
  pages={19068-19078},
  xxkeywords={Training;Costs;Video sequences;Estimation;Memory modules;Streaming media;Real-time systems},
  doi={10.1109/CVPR52733.2024.01804}}

@inproceedings{shi2023videoflow,
  author={Shi, Xiaoyu and Huang, Zhaoyang and Bian, Weikang and Li, Dasong and Zhang, Manyuan and Cheung, Ka Chun and See, Simon and Qin, Hongwei and Dai, Jifeng and Li, Hongsheng},
  booktitle={2023 IEEE/CVF International Conference on Computer Vision (ICCV)}, 
  title={{VideoFlow}: Exploiting Temporal Cues for Multi-frame Optical Flow Estimation}, 
  year={2023},
  volume={},
  number={},
  pages={12435-12446},
  xxkeywords={Bridges;Computer vision;Codes;Estimation;Bidirectional control;Benchmark testing;Iterative methods},
  doi={10.1109/ICCV51070.2023.01146}}

@inproceedings{wang2020tartanair,
  author={Wang, Wenshan and Zhu, Delong and Wang, Xiangwei and Hu, Yaoyu and Qiu, Yuheng and Wang, Chen and Hu, Yafei and Kapoor, Ashish and Scherer, Sebastian},
  booktitle={2020 IEEE/RSJ International Conference on Intelligent Robots and Systems (IROS)}, 
  title={{TartanAir}: A Dataset to Push the Limits of Visual SLAM}, 
  year={2020},
  volume={},
  number={},
  pages={4909-4916},
  xxkeywords={Visualization;Simultaneous localization and mapping;Training data;Data collection;Benchmark testing;Data models;Trajectory},
  doi={10.1109/IROS45743.2020.9341801}}

@inproceedings{mayer2016large,
  author={Mayer, Nikolaus and Ilg, Eddy and Häusser, Philip and Fischer, Philipp and Cremers, Daniel and Dosovitskiy, Alexey and Brox, Thomas},
  booktitle={2016 IEEE Conference on Computer Vision and Pattern Recognition (CVPR)}, 
  title={A Large Dataset to Train Convolutional Networks for Disparity, Optical Flow, and Scene Flow Estimation}, 
  year={2016},
  volume={},
  number={},
  pages={4040-4048},
  xxkeywords={Estimation;Three-dimensional displays;Optical imaging;Training;Cameras;Optical sensors;Optical fiber networks},
  doi={10.1109/CVPR.2016.438}}

@inproceedings{richter2017playing,
  author={Richter, Stephan R. and Hayder, Zeeshan and Koltun, Vladlen},
  booktitle={2017 IEEE International Conference on Computer Vision (ICCV)}, 
  title={Playing for Benchmarks}, 
  year={2017},
  volume={},
  number={},
  pages={2232-2241},
  xxkeywords={Benchmark testing;Semantics;Image segmentation;Games;Three-dimensional displays;Layout},
  doi={10.1109/ICCV.2017.243}}

@inproceedings{sun2025streamflow,
 author = {Sun, Shangkun and Liu, Jiaming and Li, Huaxia and Liu, Guoqing and Li, Thomas H and Gao, Wei},
 booktitle = {Advances in Neural Information Processing Systems},
 doi = {10.52202/079017-0292},
 xxeditor = {A. Globerson and L. Mackey and D. Belgrave and A. Fan and U. Paquet and J. Tomczak and C. Zhang},
 pages = {9205--9228},
 publisher = {Curran Associates, Inc.},
 title = {{StreamFlow}: Streamlined Multi-Frame Optical Flow Estimation for Video Sequences},
 volume = {37},
 year = {2024}
}

@inproceedings{weinzaepfel2022croco,
 author = {Weinzaepfel, Philippe and Leroy, Vincent and Lucas, Thomas and Br\'{e}gier, Romain and Cabon, Yohann and Arora, Vaibhav and Antsfeld, Leonid and Chidlovskii, Boris and Csurka, Gabriela and Revaud, Jerome},
 booktitle = {Advances in Neural Information Processing Systems},
 xxeditor = {S. Koyejo and S. Mohamed and A. Agarwal and D. Belgrave and K. Cho and A. Oh},
 pages = {3502--3516},
 publisher = {Curran Associates, Inc.},
 title = {{CroCo}: Self-Supervised Pre-training for 3D Vision Tasks by Cross-View Completion},
 volume = {35},
 year = {2022}
}

@inproceedings{weinzaepfel2023croco,
  author={Weinzaepfel, Philippe and Lucas, Thomas and Leroy, Vincent and Cabon, Yohann and Arora, Vaibhav and Brégier, Romain and Csurka, Gabriela and Antsfeld, Leonid and Chidlovskii, Boris and Revaud, Jérôme},
  booktitle={2023 IEEE/CVF International Conference on Computer Vision (ICCV)}, 
  title={{CroCo v2}: Improved Cross-view Completion Pre-training for Stereo Matching and Optical Flow}, 
  year={2023},
  volume={},
  number={},
  pages={17923-17934},
  doi={10.1109/ICCV51070.2023.01647}}

@inproceedings{jiang2021learning_GMA,
  author={Jiang, Shihao and Campbell, Dylan and Lu, Yao and Li, Hongdong and Hartley, Richard},
  booktitle={2021 IEEE/CVF International Conference on Computer Vision (ICCV)}, 
  title={Learning to Estimate Hidden Motions with Global Motion Aggregation}, 
  year={2021},
  volume={},
  number={},
  pages={9752-9761},
  doi={10.1109/ICCV48922.2021.00963}
}

@inproceedings{xu2022gmflow,
  author={Xu, Haofei and Zhang, Jing and Cai, Jianfei and Rezatofighi, Hamid and Tao, Dacheng},
  booktitle={2022 IEEE/CVF Conference on Computer Vision and Pattern Recognition (CVPR)}, 
  title={{GMFlow}: Learning Optical Flow via Global Matching}, 
  year={2022},
  volume={},
  number={},
  pages={8111-8120},
  doi={10.1109/CVPR52688.2022.00795}}

@inproceedings{shi2023flowformer++,
  author={Shi, Xiaoyu and Huang, Zhaoyang and Li, Dasong and Zhang, Manyuan and Cheung, Ka Chun and See, Simon and Qin, Hongwei and Dai, Jifeng and Li, Hongsheng},
  booktitle={2023 IEEE/CVF Conference on Computer Vision and Pattern Recognition (CVPR)}, 
  title={{FlowFormer++}: Masked Cost Volume Autoencoding for Pretraining Optical Flow Estimation}, 
  year={2023},
  volume={},
  number={},
  pages={1599-1610},
  doi={10.1109/CVPR52729.2023.00160}}

@inproceedings{huang2022flowformer,
author="Huang, Zhaoyang
and Shi, Xiaoyu
and Zhang, Chao
and Wang, Qiang
and Cheung, Ka Chun
and Qin, Hongwei
and Dai, Jifeng
and Li, Hongsheng",
xxeditor="Avidan, Shai
and Brostow, Gabriel
and Ciss{\'e}, Moustapha
and Farinella, Giovanni Maria
and Hassner, Tal",
title="{FlowFormer}: A Transformer Architecture for Optical Flow",
booktitle="Computer Vision -- ECCV 2022",
year="2022",
publisher="Springer Nature Switzerland",
address="Cham",
pages="668--685",
isbn="978-3-031-19790-1",
doi={10.1007/978-3-031-19790-1_40}
}

@inproceedings{saxena2023surprising,
 author = {Saxena, Saurabh and Herrmann, Charles and Hur, Junhwa and Kar, Abhishek and Norouzi, Mohammad and Sun, Deqing and Fleet, David J},
 booktitle = {Advances in Neural Information Processing Systems},
 xxeditor = {A. Oh and T. Naumann and A. Globerson and K. Saenko and M. Hardt and S. Levine},
 pages = {39443--39469},
 publisher = {Curran Associates, Inc.},
 title = {The Surprising Effectiveness of Diffusion Models for Optical Flow and Monocular Depth Estimation},
 volume = {36},
 year = {2023}
}

@article{zhao2020improved,
author={Zhao, Yuxuan
and Man, Ka Lok
and Smith, Jeremy
and Siddique, Kamran
and Guan, Sheng-Uei},
title={Improved two-stream model for human action recognition},
journal={EURASIP Journal on Image and Video Processing},
year={2020},
xxmonth={Jun},
day={17},
volume={2020},
number={1},
pages={24},
issn={1687-5281},
doi={10.1186/s13640-020-00501-x},
}

@inproceedings{huang2022real,
author="Huang, Zhewei
and Zhang, Tianyuan
and Heng, Wen
and Shi, Boxin
and Zhou, Shuchang",
xxeditor="Avidan, Shai
and Brostow, Gabriel
and Ciss{\'e}, Moustapha
and Farinella, Giovanni Maria
and Hassner, Tal",
title="Real-Time Intermediate Flow Estimation for Video Frame Interpolation",
booktitle="Computer Vision -- ECCV 2022",
year="2022",
publisher="Springer Nature Switzerland",
address="Cham",
pages="624--642",
isbn="978-3-031-19781-9",
doi={10.1007/978-3-031-19781-9_36}
}

@article{liu2020video,
author = {Liu, Xiaozhang and Liu, Hui and Lin, Yuxiu},
title = {Video frame interpolation via optical flow estimation with image inpainting},
journal = {International Journal of Intelligent Systems},
volume = {35},
number = {12},
pages = {2087-2102},
doi = {10.1002/int.22285},
year = {2020}
}

@inproceedings{xu2019quadratic,
 author = {Xu, Xiangyu and Siyao, Li and Sun, Wenxiu and Yin, Qian and Yang, Ming-Hsuan},
 booktitle = {Advances in Neural Information Processing Systems},
 xxeditor = {H. Wallach and H. Larochelle and A. Beygelzimer and F. d\textquotesingle Alch\'{e}-Buc and E. Fox and R. Garnett},
 pages = {1645--1654},
 publisher = {Curran Associates, Inc.},
 title = {Quadratic Video Interpolation},
 volume = {32},
 year = {2019}
}

@inproceedings{piergiovanni2019representation,
  author={Piergiovanni, AJ and Ryoo, Michael S.},
  booktitle={2019 IEEE/CVF Conference on Computer Vision and Pattern Recognition (CVPR)}, 
  title={Representation Flow for Action Recognition}, 
  year={2019},
  volume={},
  number={},
  pages={9937-9945},
  doi={10.1109/CVPR.2019.01018}}

@inproceedings{sun2018optical,
  author={Sun, Shuyang and Kuang, Zhanghui and Sheng, Lu and Ouyang, Wanli and Zhang, Wei},
  booktitle={2018 IEEE/CVF Conference on Computer Vision and Pattern Recognition}, 
  title={Optical Flow Guided Feature: A Fast and Robust Motion Representation for Video Action Recognition}, 
  year={2018},
  volume={},
  number={},
  pages={1390-1399},
  doi={10.1109/CVPR.2018.00151}}

@InProceedings{ilg2017flownet2,
  author={Ilg, Eddy and Mayer, Nikolaus and Saikia, Tonmoy and Keuper, Margret and Dosovitskiy, Alexey and Brox, Thomas},
  booktitle={2017 IEEE Conference on Computer Vision and Pattern Recognition (CVPR)}, 
  title={{FlowNet 2.0}: Evolution of Optical Flow Estimation with Deep Networks}, 
  year={2017},
  volume={},
  number={},
  pages={1647-1655},
  doi={10.1109/CVPR.2017.179}}

@inproceedings{sun2010secrets,
  author={Sun, Deqing and Roth, Stefan and Black, Michael J.},
  booktitle={2010 IEEE Computer Society Conference on Computer Vision and Pattern Recognition}, 
  title={Secrets of optical flow estimation and their principles}, 
  year={2010},
  volume={},
  number={},
  pages={2432-2439},
  doi={10.1109/CVPR.2010.5539939}}

@inproceedings{weinzaepfel2013deepflow,
  author={Weinzaepfel, Philippe and Revaud, Jerome and Harchaoui, Zaid and Schmid, Cordelia},
  booktitle={2013 IEEE International Conference on Computer Vision}, 
  title={{DeepFlow}: Large Displacement Optical Flow with Deep Matching}, 
  year={2013},
  volume={},
  number={},
  pages={1385-1392},
  doi={10.1109/ICCV.2013.175}}

@inproceedings{carion2020end,
author="Carion, Nicolas
and Massa, Francisco
and Synnaeve, Gabriel
and Usunier, Nicolas
and Kirillov, Alexander
and Zagoruyko, Sergey",
xxeditor="Vedaldi, Andrea
and Bischof, Horst
and Brox, Thomas
and Frahm, Jan-Michael",
title="End-to-End Object Detection with Transformers",
booktitle="Computer Vision -- ECCV 2020",
year="2020",
publisher="Springer International Publishing",
address="Cham",
pages="213--229",
isbn="978-3-030-58452-8",
doi={10.1007/978-3-030-58452-8_13}
}

@inproceedings{kirillov2023segment,
  author={Kirillov, Alexander and Mintun, Eric and Ravi, Nikhila and Mao, Hanzi and Rolland, Chloe and Gustafson, Laura and Xiao, Tete and Whitehead, Spencer and Berg, Alexander C. and Lo, Wan-Yen and Dollár, Piotr and Girshick, Ross},
  booktitle={2023 IEEE/CVF International Conference on Computer Vision (ICCV)}, 
  title={Segment Anything}, 
  year={2023},
  volume={},
  number={},
  pages={3992-4003},
  doi={10.1109/ICCV51070.2023.00371}}

@inproceedings{kerssies2025your,
  author={Kerssies, Tommie and Cavagnero, Niccolò and Hermans, Alexander and Norouzi, Narges and Averta, Giuseppe and Leibe, Bastian and Dubbelman, Gijs and De Geus, Daan},
  booktitle={2025 IEEE/CVF Conference on Computer Vision and Pattern Recognition (CVPR)}, 
  title={Your ViT is Secretly an Image Segmentation Model}, 
  year={2025},
  volume={},
  number={},
  pages={25303-25313},
  doi={10.1109/CVPR52734.2025.02356}}

@inproceedings{yang2024depth,
  author={Yang, Lihe and Kang, Bingyi and Huang, Zilong and Xu, Xiaogang and Feng, Jiashi and Zhao, Hengshuang},
  booktitle={2024 IEEE/CVF Conference on Computer Vision and Pattern Recognition (CVPR)}, 
  title={{Depth Anything}: Unleashing the Power of Large-Scale Unlabeled Data}, 
  year={2024},
  volume={},
  number={},
  pages={10371-10381},
  doi={10.1109/CVPR52733.2024.00987}}

@inproceedings{yang2024depth2,
 author = {Yang, Lihe and Kang, Bingyi and Huang, Zilong and Zhao, Zhen and Xu, Xiaogang and Feng, Jiashi and Zhao, Hengshuang},
 booktitle = {Advances in Neural Information Processing Systems},
 doi = {10.52202/079017-0688},
 xxeditor = {A. Globerson and L. Mackey and D. Belgrave and A. Fan and U. Paquet and J. Tomczak and C. Zhang},
 pages = {21875--21911},
 publisher = {Curran Associates, Inc.},
 title = {{Depth Anything V2}},
 volume = {37},
 year = {2024}
}

@inproceedings{wang2024dust3r,
  author={Wang, Shuzhe and Leroy, Vincent and Cabon, Yohann and Chidlovskii, Boris and Revaud, Jerome},
  booktitle={2024 IEEE/CVF Conference on Computer Vision and Pattern Recognition (CVPR)}, 
  title={{DUSt3R}: Geometric 3D Vision Made Easy}, 
  year={2024},
  volume={},
  number={},
  pages={20697-20709},
  doi={10.1109/CVPR52733.2024.01956}}

@inproceedings{wang2025vggt,
  author={Wang, Jianyuan and Chen, Minghao and Karaev, Nikita and Vedaldi, Andrea and Rupprecht, Christian and Novotny, David},
  booktitle={2025 IEEE/CVF Conference on Computer Vision and Pattern Recognition (CVPR)}, 
  title={{VGGT}: Visual Geometry Grounded Transformer}, 
  year={2025},
  volume={},
  number={},
  pages={5294-5306},
  doi={10.1109/CVPR52734.2025.00499}}

@inproceedings{jin2025lvsm,
 author = {Jin, Haian and Jiang, Hanwen and Tan, Hao and Zhang, Kai and Bi, Sai and Zhang, Tianyuan and Luan, Fujun and Snavely, Noah and Xu, Zexiang},
 booktitle = {International Conference on Learning Representations},
 xxeditor = {Y. Yue and A. Garg and N. Peng and F. Sha and R. Yu},
 pages = {60001--60021},
 title = {{LVSM}: A Large View Synthesis Model with Minimal 3D Inductive Bias},
 volume = {2025},
 year = {2025}
}

@inproceedings{wang2025waft,
  title={{WAFT}: Warping-Alone Field Transforms for Optical Flow},
  author={Wang, Yihan and Deng, Jia},
  booktitle={International Conference on Learning Representations},
  volume = {2026},
  year = {2026}
}

@article{wu2025study,
  title={A Study of Finetuning Video Transformers for Multi-view Geometry Tasks},
  author={Wu, Huimin and Cheng, Kwang-Ting and Lin, Stephen and Wu, Zhirong},
  journal={Proceedings of the AAAI Conference on Artificial Intelligence},
  year={2026},
  doi={10.1609/aaai.v40i13.38038}
}

@InProceedings{Bargatin_2025_ICCV,
    author    = {Bargatin, Vladislav and Chistov, Egor and Yakovenko, Alexander and Vatolin, Dmitriy},
    title     = {{MEMFOF}: High-Resolution Training for Memory-Efficient Multi-Frame Optical Flow Estimation},
    booktitle = {Proceedings of the IEEE/CVF International Conference on Computer Vision (ICCV)},
    xxmonth     = {October},
    year      = {2025},
    pages     = {8187-8196},
    doi={10.1109/ICCV51701.2025.00767}
}

@inproceedings{Bochkovskii2024:arxiv,
 author = {Bochkovskiy, Alexey and Delaunoy, Ama\"{e}l and Germain, Hugo and Santos, Marcel and Zhou, Yichao and Richter, Stephan and Koltun, Vladlen},
 booktitle = {International Conference on Learning Representations},
 xxeditor = {Y. Yue and A. Garg and N. Peng and F. Sha and R. Yu},
 pages = {75602--75637},
 title = {{Depth Pro}: Sharp Monocular Metric Depth in Less Than a Second},
 volume = {2025},
 year = {2025}
}

@inproceedings{dosovitskiy2020image,
  title={An Image is Worth 16x16 Words: Transformers for Image Recognition at Scale},
  author={Dosovitskiy, Alexey and Beyer, Lucas and Kolesnikov, Alexander and Weissenborn, Dirk and Zhai, Xiaohua and Unterthiner, Thomas and  Dehghani, Mostafa and Minderer, Matthias and Heigold, Georg and Gelly, Sylvain and Uszkoreit, Jakob and Houlsby, Neil},
  booktitle={International Conference on Learning Representations},
  year={2021}
}

@inproceedings{morimitsu2025dpflow,
  author={Morimitsu, Henrique and Zhu, Xiaobin and Cesar, Roberto M. and Ji, Xiangyang and Yin, Xu-Cheng},
  booktitle={2025 IEEE/CVF Conference on Computer Vision and Pattern Recognition (CVPR)}, 
  title={{DPFlow}: Adaptive Optical Flow Estimation with a Dual-Pyramid Framework}, 
  year={2025},
  volume={},
  number={},
  pages={17810-17820},
  doi={10.1109/CVPR52734.2025.01659}}

@inproceedings{kiefhaber2025removing,
    author    = {Kiefhaber, Simon and Roth, Stefan and Schaub-Meyer, Simone},
    title     = {Removing Cost Volumes from Optical Flow Estimators},
    booktitle = {Proceedings of the IEEE/CVF International Conference on Computer Vision (ICCV)},
    xxmonth     = {October},
    year      = {2025},
    pages     = {79-89},
    doi = {10.1109/ICCV51701.2025.00015}
}

@inproceedings{ranftl2021vision,
  author={Ranftl, René and Bochkovskiy, Alexey and Koltun, Vladlen},
  booktitle={2021 IEEE/CVF International Conference on Computer Vision (ICCV)}, 
  title={Vision Transformers for Dense Prediction}, 
  year={2021},
  volume={},
  number={},
  pages={12159-12168},
  doi={10.1109/ICCV48922.2021.01196}}

@inproceedings{liu2021swin,
  author={Liu, Ze and Lin, Yutong and Cao, Yue and Hu, Han and Wei, Yixuan and Zhang, Zheng and Lin, Stephen and Guo, Baining},
  booktitle={2021 IEEE/CVF International Conference on Computer Vision (ICCV)}, 
  title={{Swin Transformer}: Hierarchical Vision Transformer using Shifted Windows}, 
  year={2021},
  volume={},
  number={},
  pages={9992-10002},
  doi={10.1109/ICCV48922.2021.00986}}

@inproceedings{liang2021swinir,
  author={Liang, Jingyun and Cao, Jiezhang and Sun, Guolei and Zhang, Kai and Van Gool, Luc and Timofte, Radu},
  booktitle={2021 IEEE/CVF International Conference on Computer Vision Workshops (ICCVW)}, 
  title={{SwinIR}: Image Restoration Using Swin Transformer}, 
  year={2021},
  volume={},
  number={},
  pages={1833-1844},
  doi={10.1109/ICCVW54120.2021.00210}}

@inproceedings{ravisam,
 author = {Ravi, Nikhila and Gabeur, Valentin and Hu, Yuan-Ting and Hu, Ronghang and Ryali, Chaitanya and Ma, Tengyu and Khedr, Haitham and R\"{a}dle, Roman and Rolland, Chloe and Gustafson, Laura and Mintun, Eric and Pan, Junting and Alwala, Kalyan Vasudev and Carion, Nicolas and Wu, Chao-Yuan and Girshick, Ross and Dollar, Piotr and Feichtenhofer, Christoph},
 booktitle = {International Conference on Learning Representations},
 xxeditor = {Y. Yue and A. Garg and N. Peng and F. Sha and R. Yu},
 pages = {28085--28128},
 title = {{SAM 2}: Segment Anything in Images and Videos},
 volume = {2025},
 year = {2025}
}

@inproceedings{lu2023transflow,
  author={Lu, Yawen and Wang, Qifan and Ma, Siqi and Geng, Tong and Chen, Yingjie Victor and Chen, Huaijin and Liu, Dongfang},
  booktitle={2023 IEEE/CVF Conference on Computer Vision and Pattern Recognition (CVPR)}, 
  title={{TransFlow}: Transformer as Flow Learner}, 
  year={2023},
  volume={},
  number={},
  pages={18063-18073},
  doi={10.1109/CVPR52729.2023.01732}}

@inproceedings{leroy2023win,
 author = {Leroy, Vincent and Revaud, Jerome and Lucas, Thomas and Weinzaepfel, Philippe},
 booktitle = {International Conference on Learning Representations},
 xxeditor = {B. Kim and Y. Yue and S. Chaudhuri and K. Fragkiadaki and M. Khan and Y. Sun},
 pages = {48749--48767},
 title = {{Win-Win}: Training High-Resolution Vision Transformers from Two Windows},
 volume = {2024},
 year = {2024}
}

@inproceedings{liuarflow,
  title={{ARFlow}: Auto-regressive Optical Flow Estimation for Arbitrary-Length Videos via Progressive Next-Frame Forecasting},
  author={Liu, Jiuming and Liu, Mengmeng and Zhu, Siting and Zhang, Yunpeng and Li, Jiangtao and Yang, Michael Ying and Nex, Francesco and Cheng, Hao and Wang, Hesheng},
  booktitle={International Conference on Learning Representations},
  volume = {2026},
  year = {2026}
}

@article{rope2024,
title = {{RoFormer}: Enhanced transformer with Rotary Position Embedding},
journal = {Neurocomputing},
volume = {568},
pages = {127063},
year = {2024},
issn = {0925-2312},
doi = {https://doi.org/10.1016/j.neucom.2023.127063},
author = {Jianlin Su and Murtadha Ahmed and Yu Lu and Shengfeng Pan and Wen Bo and Yunfeng Liu}
}

@inproceedings{arkit2021,
 author = {Baruch, Gilad and Chen, Zhuoyuan and Dehghan, Afshin and Feigin, Yuri and Fu, Peter and Gebauer, Thomas and Kurz, Daniel and Dimry, Tal and Joffe, Brandon and Schwartz, Arik and Shulman, Elad},
 booktitle = {Proceedings of the Neural Information Processing Systems Track on Datasets and Benchmarks},
 xxeditor = {J. Vanschoren and S. Yeung},
 pages = {},
 title = {{ARKitScenes}: A Diverse Real-World Dataset For 3D Indoor Scene Understanding Using Mobile RGB-D Data},
 volume = {1},
 year = {2021}
}

@INPROCEEDINGS{megadepth2018,
  author={Li, Zhengqi and Snavely, Noah},
  booktitle={2018 IEEE/CVF Conference on Computer Vision and Pattern Recognition}, 
  title={{MegaDepth}: Learning Single-View Depth Prediction from Internet Photos}, 
  year={2018},
  volume={},
  number={},
  pages={2041-2050},
  doi={10.1109/CVPR.2018.00218}
}

@InProceedings{3dstreetview2016,
author="Zamir, Amir R.
and Wekel, Tilman
and Agrawal, Pulkit
and Wei, Colin
and Malik, Jitendra
and Savarese, Silvio",
xxeditor="Leibe, Bastian
and Matas, Jiri
and Sebe, Nicu
and Welling, Max",
title="Generic 3D Representation via Pose Estimation and Matching",
booktitle="Computer Vision -- ECCV 2016",
year="2016",
publisher="Springer International Publishing",
address="Cham",
pages="535--553",
isbn="978-3-319-46487-9",
doi={10.1007/978-3-319-46487-9_33}
}

@INPROCEEDINGS{hd1k2016,
  author={Kondermann, Daniel and Nair, Rahul and Honauer, Katrin and Krispin, Karsten and Andrulis, Jonas and Brock, Alexander and Güssefeld, Burkhard and Rahimimoghaddam, Mohsen and Hofmann, Sabine and Brenner, Claus and Jähne, Bernd},
  booktitle={2016 IEEE Conference on Computer Vision and Pattern Recognition Workshops (CVPRW)}, 
  title={{The HCI Benchmark Suite}: Stereo and Flow Ground Truth with Uncertainties for Urban Autonomous Driving}, 
  year={2016},
  volume={},
  number={},
  pages={19-28},
  doi={10.1109/CVPRW.2016.10}
}

@InProceedings{Zhai_2022_CVPR,
    author    = {Zhai, Xiaohua and Kolesnikov, Alexander and Houlsby, Neil and Beyer, Lucas},
    title     = {Scaling Vision Transformers},
    booktitle = {Proceedings of the IEEE/CVF Conference on Computer Vision and Pattern Recognition (CVPR)},
    xxmonth     = {June},
    year      = {2022},
    pages     = {12104-12113},
    doi = {10.1109/CVPR52688.2022.01179}
}

\clearpage
\setcounter{page}{1}
\appendix
\title{FreeFlow: A Bias-free Hierarchical \\Transformer for Optical Flow Estimation}
\author{Supplementary Material}
\institute{}
\maketitle

This supplementary provides additional details and context on training and evaluation protocols, metric and loss definitions, and extended qualitative and ablation results, and is organized as follows:
\begin{itemize}
    \item \cref{sec:abl:def} formally defines evaluation metrics and the loss function;
    \item \cref{sec:abl:crocopro} discusses Dense Feature Fusion and the effectiveness of FreeFlow;
    \item \cref{sec:abl:add-res} provides additional ablations and results;
    \item \cref{sec:abl:qual} shows more qualitative examples.
\end{itemize}

\section{Definitions}
\label{sec:abl:def}

In the following sections, $\mu_{\text{gt}}(u, v)$ is the target flow vector at position $(u, v)$, $\mu$ is the predicted flow vector at position $(u, v)$, $N$ is the number of valid pixels in the target flow field, and $[\cdot]$ is the Iverson bracket. Sums of the form $\sum_{u,v}$ are calculated only over valid pixels.

\subsection{Metrics}

\subsubsection{Endpoint Error.}
Endpoint error (EPE) is defined as:
\begin{align}
    \textbf{EPE} &= \frac{1}{N} \sum_{u,v} \| \mu_{\text{gt}}(u,v) - \mu(u,v) \|_2.
\end{align}
and ranges from $+\infty$ at worst to $0$ at best. It is adopted as the main metric for the Sintel benchmark.

\subsubsection{One-pixel Outlier Rate.}
The 1-pixel outlier rate (1px) is defined as:
\begin{align}
    \textbf{1px} &= \frac{100}{N} \sum_{u,v} \left[\| \mu_{\text{gt}}(u,v) - \mu(u,v) \|_2 > 1\right].
\end{align}
and ranges from $100$ at worst to $0$ at best. It is adopted as the main metric for the Spring benchmark.

\subsubsection{Fl-all Outliers Metric.}
The Fl-all outliers metric (Fl-all score) is defined as:
\begin{align}
    \textbf{Fl-all} &= \frac{100}{N} \sum_{u,v} \left[\| \mu_{\text{gt}}(u,v) - \mu(u,v) \|_2 > \max(0.05 \cdot \mu_{\text{gt}}(u,v), 3) \right],
\end{align}
and ranges from $100$ at worst to $0$ at best. Is is adopted as the main metric for the KITTI-15 benchmark due to the noisy nature of the collected real-world data.

\subsubsection{Weighted Area Under the Curve.} The weighted area under the curve (WAUC) is formally defined as:

\begin{align}
    \textbf{WAUC} &= \frac{2}{5}\int_0^5 \left( \frac{100}{N} \sum_{u,v} [\| \mu_{\text{gt}}(u,v) - \mu(u,v) \|_2 \le x] \right) \cdot \frac{5 - x}{5}\, dx,
\end{align}
and ranges from $0$ at worst to $100$ at best. In practice, this integral is usually approximated with $100$ bins. WAUC can be also viewed as a generalization of 1px score.

\subsection{Mixture-of-Laplace Loss}
\label{sec:sup:mol}

For a single flow vector coordinate, the Mixture-of-Laplace (MoL) in SEA-RAFT is defined as:
\begin{align}
    \text{MixLap}(\mu_{gt}; \alpha, \beta, \mu) &= -\log \left( \frac{\alpha}{2} \cdot e^{-|\mu_{gt}-\mu|} + \frac{1-\alpha}{2e^{\beta}} \cdot e^{-\frac{|\mu_{gt}-\mu|}{e^{\beta}}} \right),
\end{align}
where $\mu_{\text{gt}}$ is the target flow coordinate, $\mu$ is the predicted flow coordinate, $\alpha$ is the predicted mixing coefficient, and $\beta$ is the predicted scale parameter, clamped to the range $[0, 10]$. In practice and as is used in the official implementation, $\alpha$ and $1 - \alpha$ are calculated as the softmax of two outputs $\alpha_1$ and $\alpha_2$. The final MoL loss is then defined as:

\begin{align}
    \mathcal{L}_{MoL} = \frac{1}{2N} \sum_{u,v} \sum_{d \in \{x,y\}} \text{MixLap}\bigl(\mu_{\text{gt}}(u,v)_d; \alpha(u,v), \beta(u,v), \mu(u,v)_d \bigr).
\end{align}

\begin{figure}[t]
    \centering
    \begin{subfigure}[t]{0.49\linewidth}
        \centering
        \includegraphics[width=\linewidth]{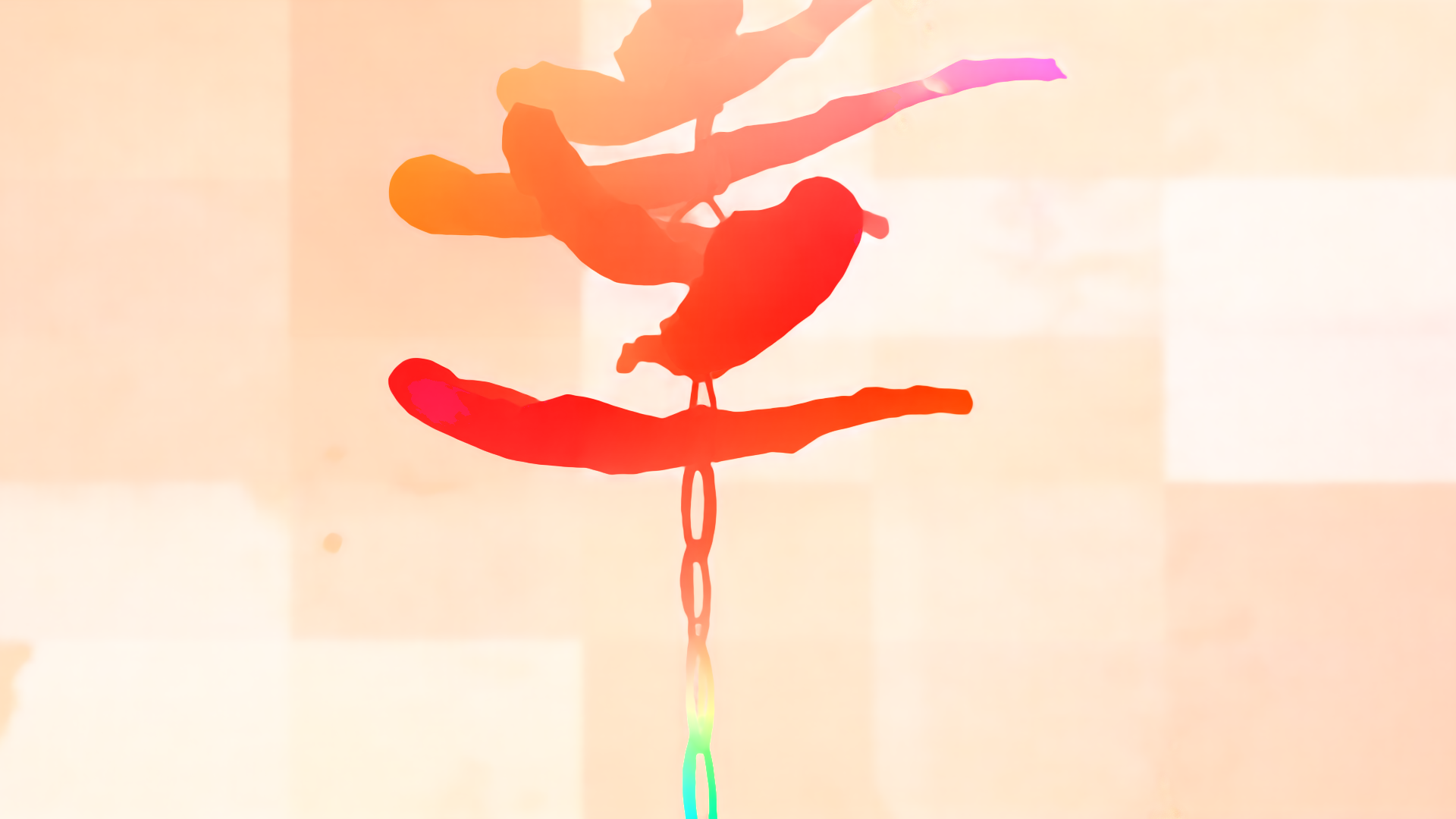}
        \frame{\includegraphics[width=0.997\linewidth]{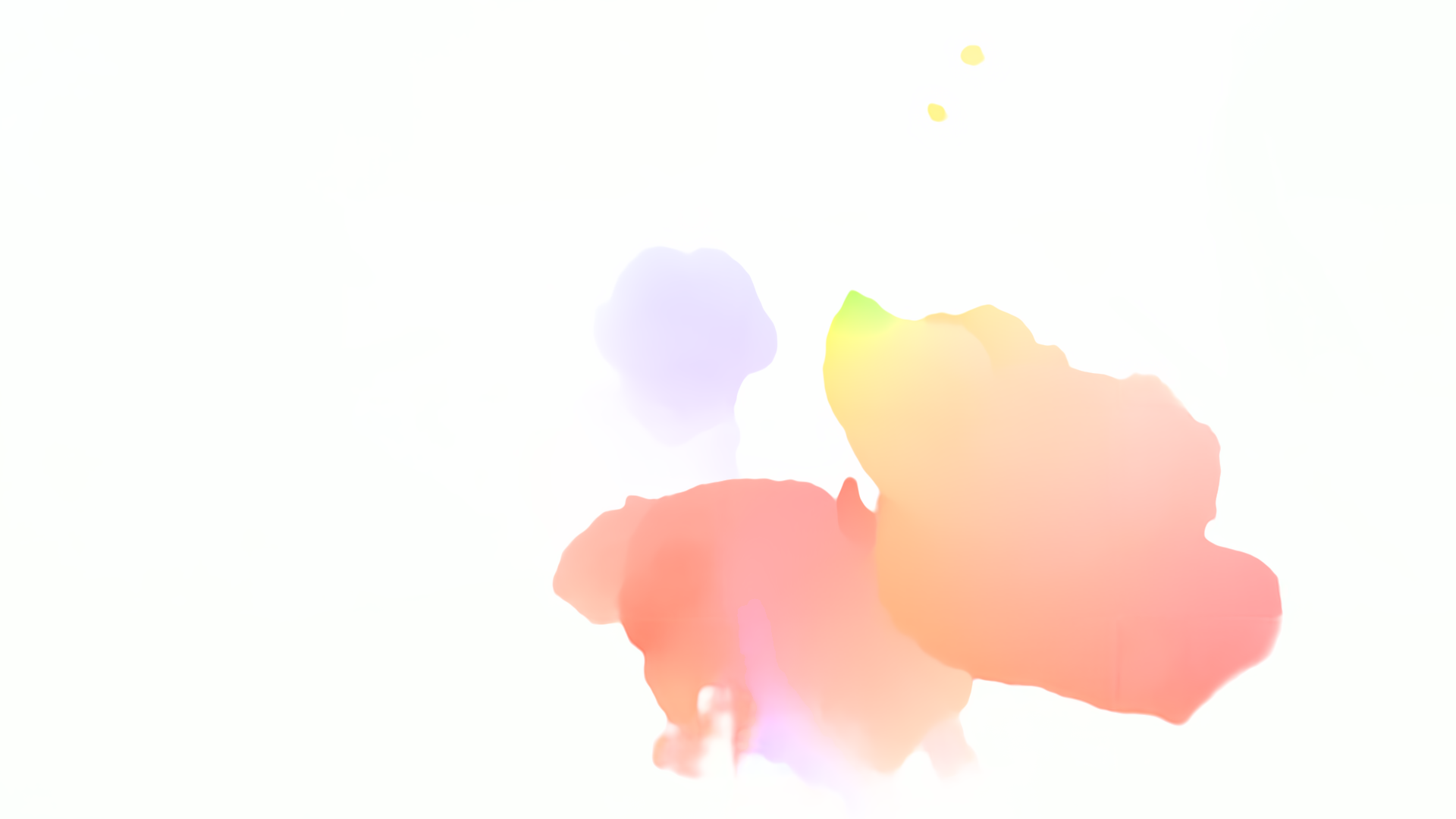}}
        \caption{CroCo-Pro (Late Feature Fusion)}
        \label{fig:sup:provsfree:pro}
    \end{subfigure}
    \begin{subfigure}[t]{0.49\linewidth}
        \centering
        \includegraphics[width=\linewidth]{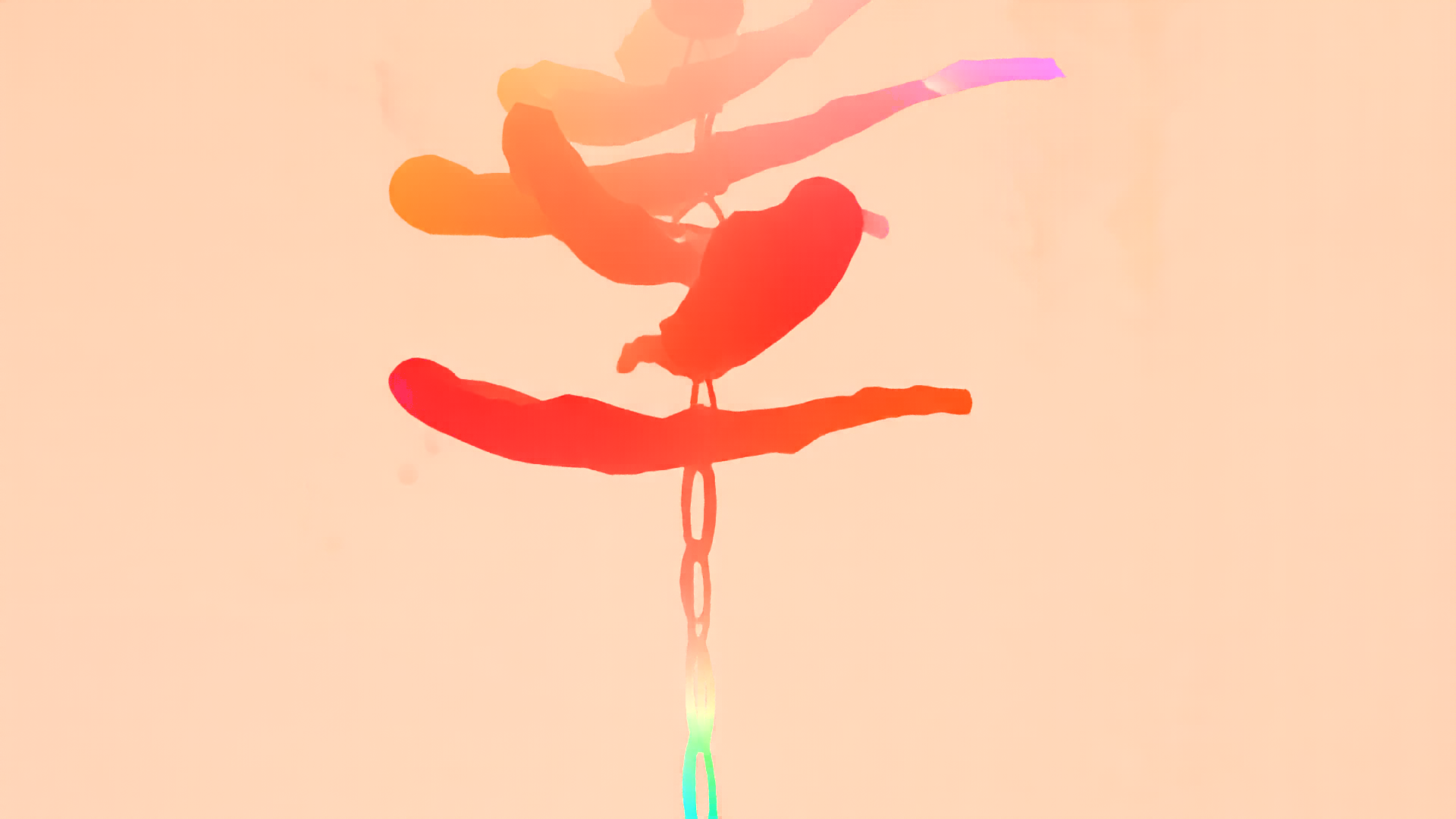}
        \frame{\includegraphics[width=0.997\linewidth]{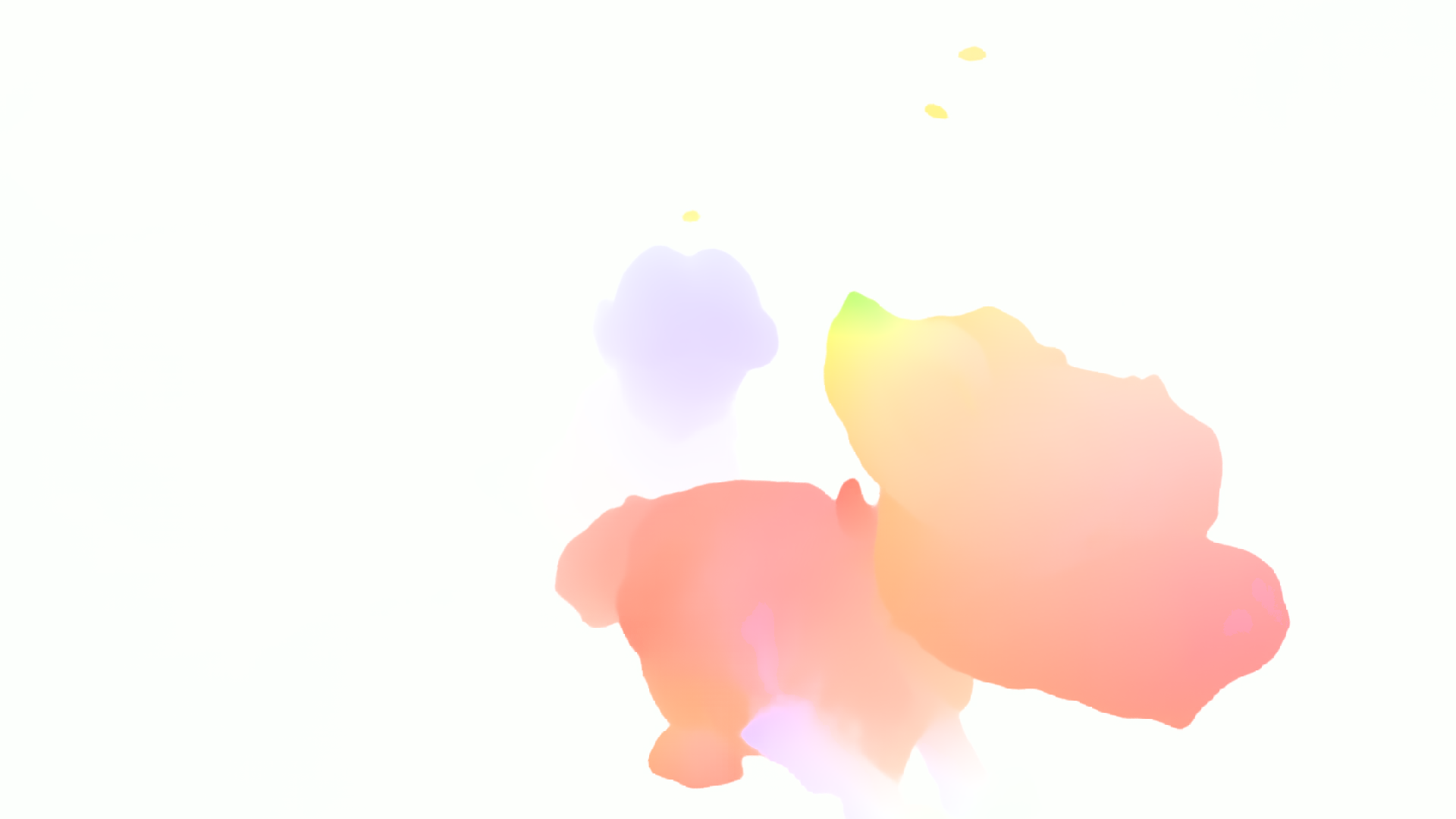}}
        \caption{FreeFlow-L (Dense Feature Fusion)}
        \label{fig:sup:provsfree:free}
    \end{subfigure}
    \caption{\textbf{Feature fusion ablation.} We compare a DepthPro-style \emph{Late Feature Fusion} baseline (CroCo backbone with multi-scale decoding) to FreeFlow with \emph{Dense Feature Fusion}. Late fusion does not reliably propagate information between scales and tiles: the model tends to rely on the finest-level tiles and under-utilize coarse-scale features, leading to globally inconsistent flow despite sharp local detail (left). Dense fusion exchanges features throughout the network, yielding predictions that are both locally detailed and globally consistent (right).}
    \label{fig:sup:provsfree}
\end{figure}

\section{Architecture Discussion}
\label{sec:abl:crocopro}

To isolate the role of feature fusion, we implemented a DepthPro-style \emph{Late Feature Fusion} baseline by replacing the monocular ViT backbone with a pretrained CroCo encoder--decoder and fusing pyramid features in a DPT-decoder (which we call CroCo-Pro). As shown in \cref{fig:sup:provsfree:pro}, this design does not reliably propagate information across scales and tile boundaries: coarse-scale cues are weakly utilized and have limited influence on the final full-resolution prediction. The resulting flow can preserve fine local structure, yet exhibits reduced global coherence, especially when displacements span multiple tiles or when texture is ambiguous.

FreeFlow addresses this limitation by exchanging features throughout the network rather than only at the final decoding stage (\cref{fig:sup:provsfree:free}). Shifted-window blocks provide repeated cross-tile communication, while reduced-resolution global attention injects long-range context, allowing coarse and fine signals to reinforce each other during decoding. Beyond accuracy, this design is practical at high resolution: all interaction mechanisms are expressed via standard attention operators (windowed, shifted-window, and global attention on a downsampled map), whose token counts are balanced so that their computational cost is comparable. As a result, FreeFlow directly benefits from off-the-shelf efficient attention kernels that have been heavily optimized in recent years, whereas flow-specific modules such as correlation volume indexing, feature warping, or iterative update pipelines require task-specific engineering to reach similar efficiency.

\section{Additional results}
\label{sec:abl:add-res}

\subsection{Training Details}
We use AdamW with $\beta_1\!=\!0.9$ and $\beta_2\!=\!0.95$. For pretraining, we adopt a linear warmup followed by a cosine learning-rate decay. For optical flow finetuning, we use a linear warmup followed by a linear decay schedule, which is standard in optical flow training. We train with the Mixture-of-Laplace (\cref{sec:sup:mol}) loss from SEA-RAFT.

\begin{table}[t]
    \caption{\textbf{Patch size ablation.} We compare FreeFlow-S with a $16\!\times\!16$ variant at matched 1080p inference time. To account for the larger image area represented by each token, we increase the variant's width and number of attention heads. The base model still achieves better 1px accuracy with much lower memory and parameter cost.}
    \label{tab:sup:patch-size}
    \centering
    \resizebox{1.0\linewidth}{!}{
\begin{tabular}{c c c c c c c c c}
\toprule
\multirow{2}[2]{*}{\makecell[c]{Patch\\size}} &
\multirow{2}[2]{*}{\makecell[c]{Pre-train\\crop}} &
\multirow{2}[2]{*}{\makecell[c]{Mask.\\ratio}} &
\multirow{2}[2]{*}{Width} &
\multicolumn{2}{c}{Inf. Cost (1080p)}
& \multirow{2}[2]{*}{Params (M)}
& \multicolumn{2}{c}{Spring (sub-val)} \\
\cmidrule(lr){5-6} 
\cmidrule(lr){8-9}
 & & & & Memory (GB) & Time (ms) & & 1px $\downarrow$ & EPE $\downarrow$ \\
\midrule
\rowcolor{lightgray}
8$\times$8 & [224, 224] & 0.95 & 256 & \textbf{1.02} & 144 & \textbf{34.58} & 0.181 & \textbf{0.709} \\
16$\times$16 & [256, 256] & 0.9 & 768 & 2.91 & \textbf{120} & 278.88 & \textbf{0.174} & 0.798 \\
\bottomrule
\end{tabular}
}
\end{table}

\subsection{Patch Size Ablation} 
We investigate the role of the patch size on our method's performance. As the base $8\!\times\!8$ model, we take FreeFlow-S. For the $16\!\times\!16$ model, to make for a fair comparision, we increase the width from $256$ to $768$ and the number of attention heads from $4$ to $12$, resulting in similar execution time.

Tab.~\ref{tab:sup:patch-size} shows that the $16\!\times\!16$ variant is only marginally better in EPE (0.174 vs.\ 0.181), while being worse in 1px (0.798 vs.\ 0.709) and substantially more expensive in parameters and memory (279M and 2.91\,GB vs.\ 35M and 1.02\,GB). Overall, this supports using smaller patches in our setting, as they provide comparable accuracy at a substantially lower memory and parameter cost.

\subsection{Comparison to Vanilla ViT}

To study the performance of the proposed architecture separately from the used training procedure, we trained a variant in which FreeFlow encoder/decoder modules are substituted for vanilla ViTs. This model closely resembles Croco-Flow's and Win-Win's designs (except for the DPT-head used in the postprocessing stage). The results are presented in ~\cref{tab:sup:vit-warp} (rows 1--3). 
Even within a larger computational budget (270--410\,ms vs 131\,ms), the ViT-based model results in a larger prediction error, while the quality difference between the ViT and FreeFlow models with similar parameter counts is negligible (with the ViT model being almost 6 times slower). This confirms that the proposed Local-Global attention design, not the training recipe alone, improves on the quality-performance tradeoff of a standard ViT-based approach.

\subsection{Injecting Biases back into FreeFlow}

We modified FreeFlow to include explicit optical flow biases, testing a GeoViT-like iterative warping procedure as it is straightforward to integrate and currently among the most effective iterative transformer-based approaches. We do not change the model during pretraining, only introducing warping and the recursive module (the same ConvGRU as in GeoViT) during the optical flow fine-tuning stage. Results are provided in ~\cref{tab:sup:vit-warp} (rows 6--8). The GeoViT-like variant shows slightly better prediction quality within the same parameter budget in some configurations, but at the cost of slightly slower inference due to its iterative nature. This is consistent with expectations: the inductive bias offloads task-specific knowledge into an explicit operator, increasing the effective capacity of the model. This further illustrates that flow-specific biases are not necessary to reach SOTA performance, but can be added to FreeFlow to improve accuracy at the cost of speed and architectural generality.

\begin{table}[t]
    \caption{Comparison with close alternative approaches, results on Spring (sub-val) after Spring (sub-train). All models have a width of 256 with 4 attention heads (except for FreeFlow-S+, which has 8), used a masking ratio of 0.95 during pre-training, and followed the training procedure described above. The "iters" column indicates the number of iterative refinements used in GeoViT warping (during both training and inference). "+" splits time into the transformer and the postprocessing head parts.}
    \label{tab:sup:vit-warp}
    \centering
    \begin{tabular}{l c c c c c c c}
\toprule
\multirow{2}[2]{*}{Type} & \multirow{2}[2]{*}{Layers} & \multirow{2}[2]{*}{Iters} & \multirow{2}[2]{*}{\makecell[c]{Time (ms)\\(1080p)}} & \multirow{2}[2]{*}{\makecell[c]{Params\\(M)}} & \multicolumn{2}{c}{Spring (sub-val)} \\
\cmidrule(lr){6-7}
& & & & & 1px$\downarrow$ & EPE$\downarrow$ \\
\midrule
\multirow{3}{*}{ViT}
 & 4 & --- & 278+13 & 15.36 & 1.021 & 0.217 \\
 & 6 & --- & 415+13 & 19.05 & 0.809 & 0.216 \\
 & 12 & --- & 829+13 & 30.11 & 0.698 & 0.192 \\
\cdashlinelr{1-7}
FreeFlow-S
 & 4 & --- & 131+13 & 34.58 & 0.709 & 0.181 \\
FreeFlow-S+
 & 8 & --- & 256+13 & 60.90 & 0.624 & 0.157 \\
\cdashlinelr{1-7}
\multirow{3}[1]{*}{\makecell[c]{
FreeFlow-S\\with GeoViT\\warping
}}
 & 4 & 1 & 131+40 & 33.12 & 0.660 & 0.165 \\
 & 2 & 2 & 131+62 & 19.96 & 0.719 & 0.171 \\
\cdashlinelr{2-7}
 & 4 & 2 & 262+53 & 33.12 & 0.583 & 0.148 \\
\bottomrule
\end{tabular}
\end{table}


\begin{table}[t]
    \caption{\textbf{Zero-shot comparison}. We report zero-shot evaluation results on the Sintel and KITTI-15 training sets. By default, all methods are trained or fine-tuned for optical flow estimation on (FlyingChairs +) FlyingThings3D, with the "TA" column indicating that TartanAir was additionally used. Method biases abbreviations: IR: Iterative Refinement, CV: Correlation Volume, W: Warping. The pre-train column indicates whether some part of the method was not trained from scratch, size is reported as the total number of frames in the dataset (for CroCo, this is double the number of image pairs, for Kinetics, this is the total number of frames in all videos). "MF" indicates that the method used multiple frames (three or more) to generate it's submissions.}
    \label{tab:sup:zero-shot}
    \centering
    \newcommand{\yay}{\cellcolor{red!25}$\checkmark$\xspace}
\newcommand{\nay}{\cellcolor{green!25}$\times$\xspace}
\newcommand{\nayw}{\cellcolor{white!25}$\times$\xspace}

\resizebox{1.0\linewidth}{!}{
\begin{tabular}{l ccc cc c cc cc}
\toprule
\multirow{2}[2]{*}{Method} & \multicolumn{3}{c}{Biases} & \multicolumn{2}{c}{Pre-train} & \multirow{2}[2]{*}{TA} & \multicolumn{2}{c}{Sintel (train)} & \multicolumn{2}{c}{KITTI-15 (train)} \\
\cmidrule(lr){2-4}\cmidrule(lr){5-6}\cmidrule(lr){8-9}\cmidrule(lr){10-11}
 & IR & CV & W & Name & Size & & Clean$\downarrow$ & Final$\downarrow$ & Fl-epe$\downarrow$ & Fl-all$\downarrow$ \\
\midrule
RAFT & \yay & \yay & \nay & --- & --- & $\times$ & 1.43 & 2.71 & 5.04 & 17.4 \\
GMA & \yay & \yay & \nay & --- & --- & $\times$ & 1.30 & 2.74 & 4.69 & 17.1 \\
FlowFormer & \yay & \yay & \nay & ImageNet-1K & 1.3M & $\times$ & 1.01 & 2.40 & 4.09 & 14.7 \\
SEA-RAFT (S) & \yay & \yay & \nay & ImageNet-1K & 1.3M & $\checkmark$ & 1.27 & 3.74 & 4.43 & 15.1 \\
SEA-RAFT (M) & \yay & \yay & \nay & ImageNet-1K& 1.3M & $\times$ & 1.21 & 4.04 & 4.29 & 14.2 \\
SEA-RAFT (L) & \yay & \yay & \nay & ImageNet-1K& 1.3M & $\times$ & 1.19 & 4.11 & 3.62 & 12.9 \\
DPFlow & \yay & \yay & \nay & --- & --- & $\times$ & 1.02 & 2.26 & 3.37 & 11.1 \\
VideoFlow-BOF\textsuperscript{(MF)} & \yay & \yay & \nay & ImageNet-1K& 1.3M & $\times$ & 1.03 & 2.19 & 3.96 & 15.3 \\
VideoFlow-MOF\textsuperscript{(MF)} & \yay & \yay & \nay & ImageNet-1K& 1.3M & $\times$ & 1.18 & 2.56 & 3.89 & 14.2 \\
MemFlow\textsuperscript{(MF)} & \yay & \yay & \nay & --- & --- & $\times$ & 0.93 & 2.08 & 3.88 & 13.7 \\
MemFlow-T\textsuperscript{(MF)} & \yay & \yay & \nay & ImageNet-1K& 1.3M & $\times$ & \underline{0.85} & \underline{2.06} & 3.38 & 12.8 \\
StreamFlow\textsuperscript{(MF)} & \yay & \yay & \nay & ImageNet-1K& 1.3M & $\times$ & \textit{0.87} & 2.11 & 3.85 & 12.6 \\
MEMFOF\textsuperscript{(MF)} & \yay & \yay & \nay & ImageNet-1K& 1.3M & $\times$ & 1.10 & 2.70 & 3.31& \textit{10.1} \\
MEMFOF\textsuperscript{(MF)} & \yay & \yay & \nay & ImageNet-1K& 1.3M & $\checkmark$ & 1.20 & 3.91 & \underline{2.93} & \underline{9.9} \\
ARFlow\textsuperscript{(MF)} & \yay & \yay & \nay & ImageNet-1K& 1.3M & $\checkmark$ & 0.88 & \textit{2.07} & \textbf{2.86} & \textbf{9.2} \\
WAFT-Twins-a2 & \yay & \nay & \yay & ImageNet-1K& 1.3M & $\checkmark$ & 1.02 & 2.46 & \textit{2.98} & \underline{9.9} \\
WAFT-DAv2-a2 & \yay & \nay & \yay & DAv2 & 63M & $\checkmark$ & 1.01 & 2.49 & 3.28 & 10.9 \\
WAFT-DINOv3-a2 & \yay & \nay & \yay & LVD-1689M & 1.7B & $\checkmark$ & 1.28 & 2.56 & 3.49 & 12.9 \\
GeoViT & \yay & \nay & \yay & Kinetics-400 & 59M & $\times$ & \textbf{0.69} & \textbf{1.78} & 3.15 & 11.5 \\
CroCo-Flow & \nay & \nay & \nay & CroCo v2 & 15M & $\times$ & 1.28 & 2.58 & --- & --- \\
FreeFlow-S (ours) & \nay & \nay & \nay & CroCo v2 & 7.4M & $\checkmark$ & 0.91 & 3.16 & 3.41 & 10.4 \\
FreeFlow-M (ours) & \nay & \nay & \nay & CroCo v2 & 7.4M & $\checkmark$ & 1.01 & 3.12 & 5.89 & 14.6 \\
FreeFlow-L (ours) & \nay & \nay & \nay & CroCo v2 & 7.4M & $\checkmark$ & 1.04 & 2.30 & 4.77 & 12.9 \\
\bottomrule
\end{tabular}
}
\end{table}

\begin{figure}[t]
  \centering
  \includegraphics[width=0.99\linewidth]{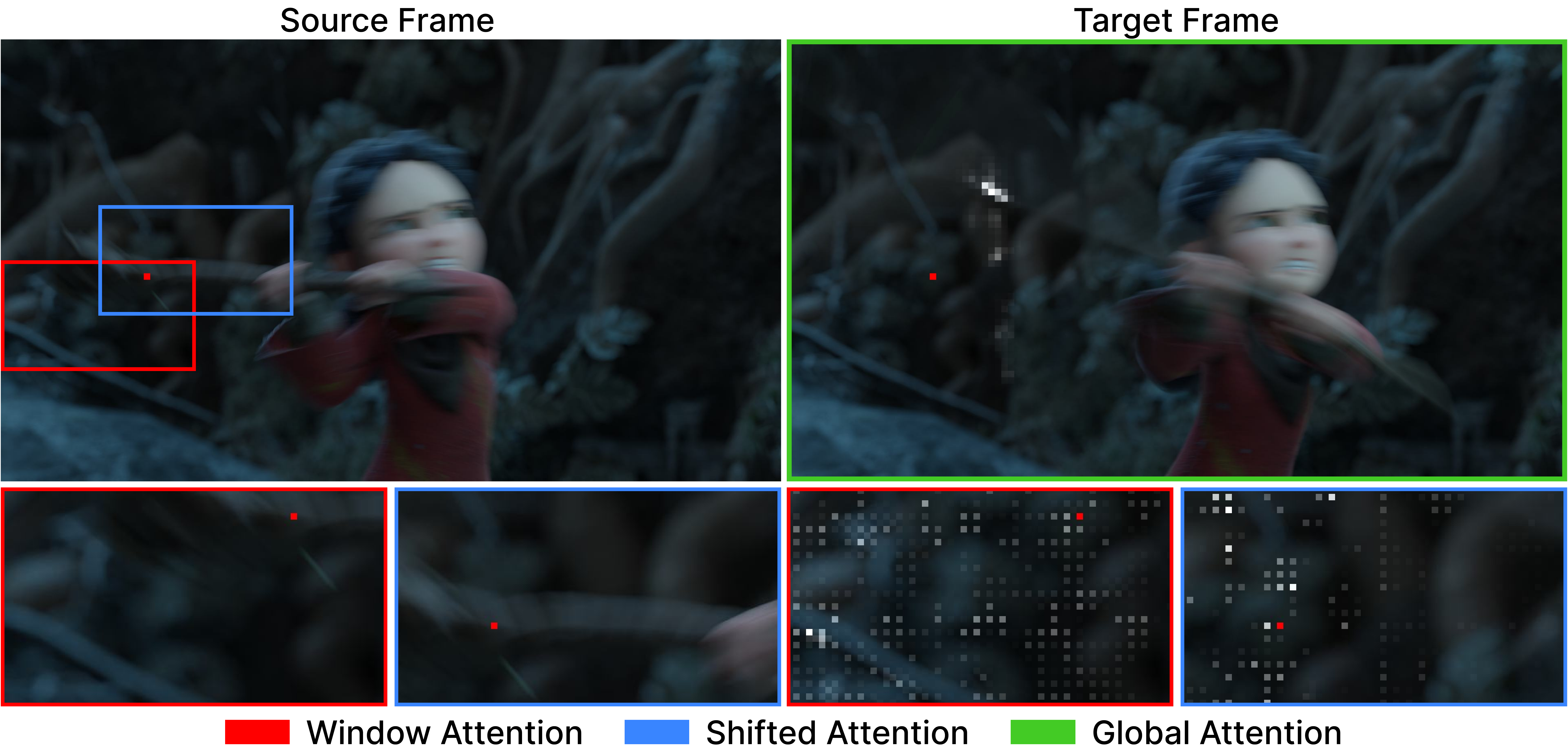}
   \caption{Visualization of the softmax logits for different parts of the proposed local-global attention. The red point represents the query token.}
   \label{fig:onecol}
\end{figure}

\subsection{Zero-shot Performance}
We evaluate zero-shot transfer on the Sintel and KITTI training sets by finetuning only on TartanAir (TA) and FlyingThings3D. Specifically, we remove the Sintel, KITTI, and HD1K portions from the TaTSKH / TaTSKH-hq stages and report performance after completing the ``hq'' stage (\cref{tab:sup:zero-shot}).

FreeFlow attains reasonable zero-shot performance on Sintel and KITTI under TA+Things finetuning, but does not improve monotonically with model size. This behavior is expected for a bias-free model in a data-limited regime: transfer is primarily constrained by the motion and appearance coverage of the available training signal, and increasing capacity alone does not guarantee gains.

This trend is reflected by the methods that use broader pretraining datasets. GeoViT, pretrained on the large Kinetics-400 video dataset, achieves the strongest zero-shot performance on Sintel, consistent with exposure to substantially more varied motion. In contrast, CroCo-Flow, pretrained only with CroCo-style data and without the use of TA during training, exhibits weaker transfer. Finally, the WAFT variants suggest that large-scale \emph{monocular} pretraining alone is not always sufficient for zero-shot optical flow: despite substantially larger pretraining datasets, their transfer does not match methods pretrained on binocular or video data, indicating the importance of multi-view and motion-centric pretraining signals.

\subsection{Additional Model Analysis}

We show visualizations for different attention types in Fig.~\ref{fig:onecol}: global attention helps FreeFlow capture large displacements, while local attention processes small shifts. On shifted-window attention in the decoder, the window partition is shifted identically in both frames so that corresponding regions remain co-located across the pair.

\section{Additional Qualitative Comparisons}
\label{sec:abl:qual}
\begin{figure}[t]
\centering
\includegraphics[width=.99\textwidth]{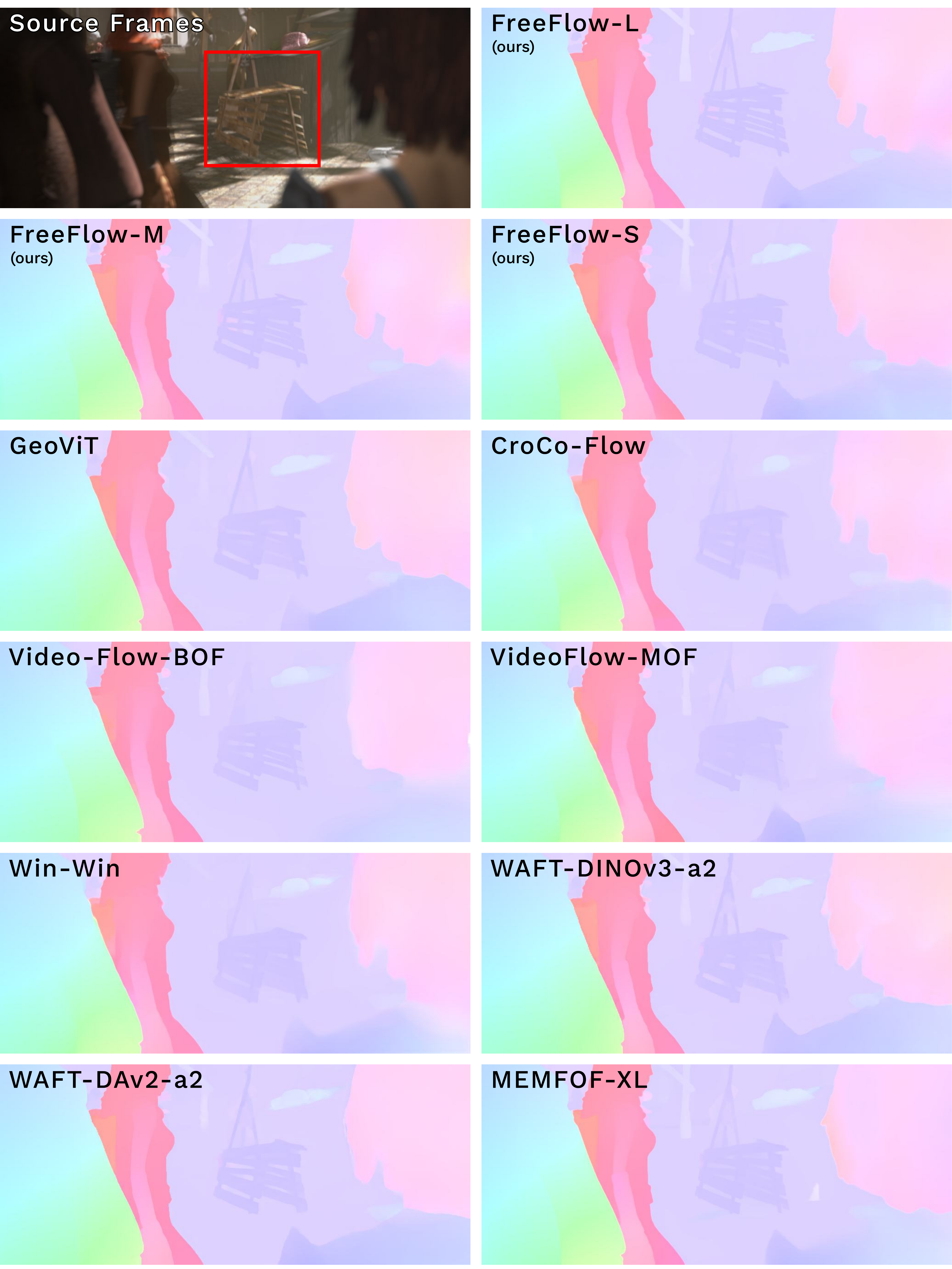}\\
\caption{\textbf{Additional qualitative samples on Sintel.} Across the board FreeFlow models produce the sharpest details and effectively separate the wooden structure from the background. Images are sourced from the official leaderboard webpages. Viewer is advised to zoom in.}
\label{fig:sup:sintel_viz}
\end{figure}

\begin{figure}[t]
\centering
\includegraphics[width=.9\textwidth]{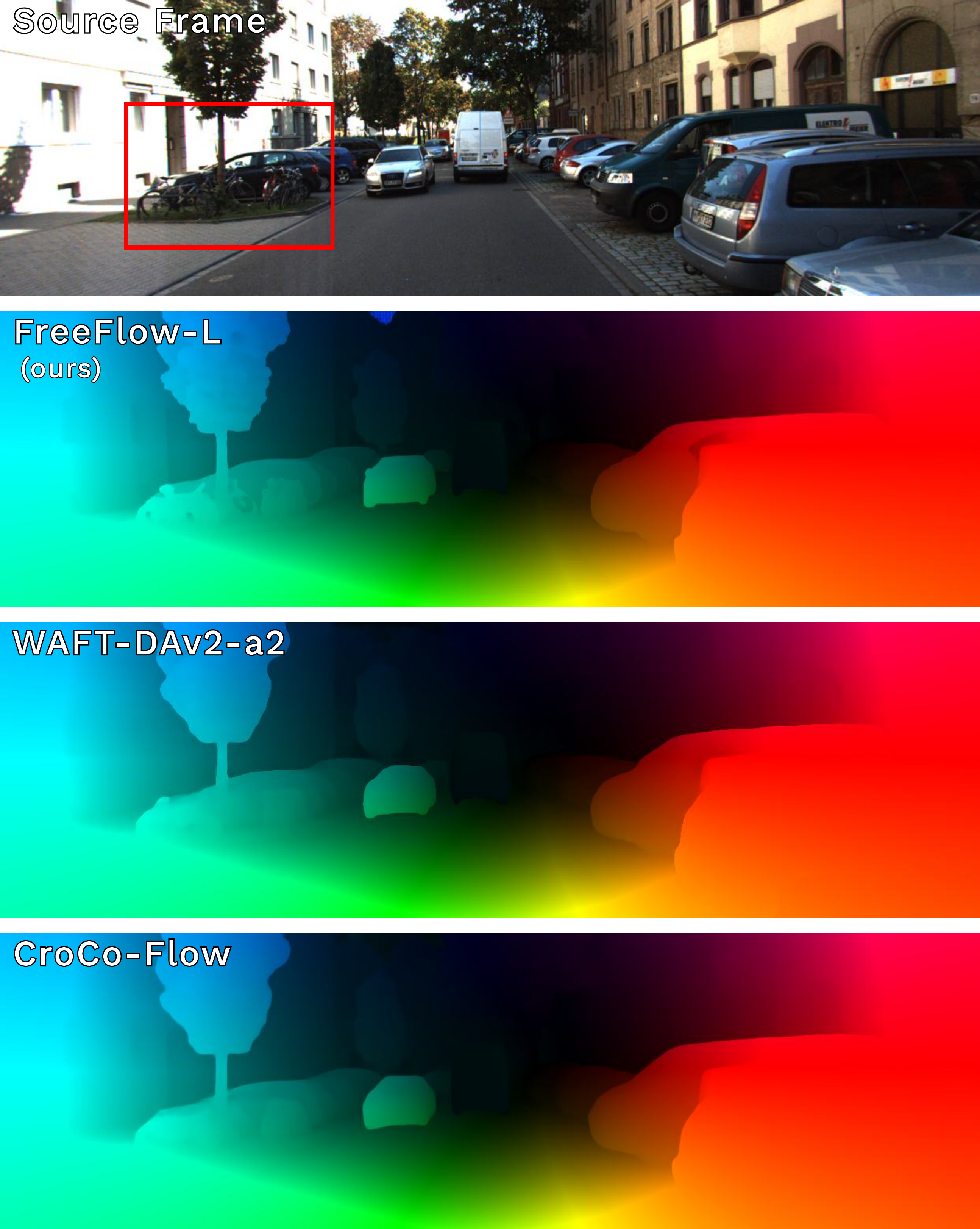}\\
\caption{\textbf{Additional qualitative samples on KITTI-15.} FreeFlow has significantly more details on complex objects such as the wheels of bicycles, side-view mirrors of cars and tree foliage. Images are sourced from the official leaderboard webpages. Viewer is advised to zoom in.}
\label{fig:sup:kitti_viz}
\end{figure}

We provide additional qualitative samples for Sintel (\cref{fig:sup:sintel_viz}) and KITTI-15 (\cref{fig:sup:kitti_viz}) datasets.

\end{document}